\documentclass[preprint,12pt]{elsarticle}

\usepackage{amsmath,amssymb}
\usepackage{booktabs}
\usepackage{multirow}
\usepackage{graphicx}
\graphicspath{{figures/}}
\usepackage{xcolor}
\usepackage{hyperref}
\usepackage{algorithm}
\usepackage{algpseudocode}
\usepackage{siunitx}
\usepackage{subcaption}
\usepackage{float}        % provides [H] for exact placement
\usepackage{placeins}     % provides \FloatBarrier
\usepackage{makecell}     % provides \makecell for line-wrapped table headers

\journal{Geoenergy Science and Engineering (preprint, arXiv submission)}

\newcommand{\wisemodel}{\textsc{WISE-FM}}
\newcommand{\film}{\textsc{FiLM}}

\newcommand{\boldentry}[1]{\textbf{#1}}

\begin{document}

\begin{frontmatter}

\title{WISE-FM:Operation-Aware, Engineering-Informed Foundation Model for Multi-Task Well Design}

\author[ntnu]{Carine de Menezes Rebello}
\author[petrobras]{Anderson Rapello dos Santos}
\author[ntnu]{Idelfonso B.R.\ Nogueira\corref{cor1}}
\ead{idelfonso.nogueira@ntnu.no}
\cortext[cor1]{Corresponding author.}

\affiliation[ntnu]{organization={Department of Chemical Engineering, Norwegian University of Science and Technology (NTNU)},
  city={Trondheim},
  country={Norway}}

\affiliation[petrobras]{organization={Petróleo Brasileiro S.A.\ (PETROBRAS), Wells/Well Engineering},
  city={Rio de Janeiro},
  country={Brazil}}

\begin{abstract}
Deploying machine learning models across diverse well portfolios requires generalisation to wells with design parameters outside the training distribution. Current data-driven approaches to virtual flow metering (VFM) and bottomhole estimation typically treat each well independently or ignore the influence of well design on operational behaviour. We present \wisemodel{} (Well Intelligence and Systems Engineering Foundation Model), a design-aware, physics-informed multi-task model that integrates three complementary mechanisms: Feature-wise Linear Modulation (\film{}) and cross-modal attention to condition operational embeddings on well design parameters; multi-task learning for simultaneous prediction of flow rates, bottomhole conditions, and flow regime classification; and structural mass conservation with soft physics constraints derived from well engineering principles. Evaluation on the ManyWells benchmark (2000 simulated wells, $10^6$ data points) demonstrates that design-aware models reduce VFM prediction error by up to $13\times$ compared to design-unaware baselines, and that physics constraints reduce negative flow predictions by 65\%. Flow regime classification achieves 97.7\% bottomhole accuracy, providing continuous well integrity monitoring without additional sensors. The methodology transfers to real operational data from five Equinor Volve producers (oil rate $R^2 = 0.89$, bottomhole pressure $R^2 = 0.98$, water rate $R^2 = 0.97$). The trained model additionally serves as a fast surrogate for integrity-aware well design optimisation over a 24-dimensional design space, with more than $1000\times$ speedup over drift-flux simulations. These results demonstrate that design awareness, physics enforcement, and multi-task learning are essential and complementary ingredients for foundation models intended to operate across large well portfolios.
\end{abstract}


\end{frontmatter}

%=============================================================================
\section{Introduction}
\label{sec:intro}
%=============================================================================

The petroleum industry operates hundreds of thousands of wells worldwide, each with a unique combination of design parameters (tubing diameter, well depth, completion geometry, reservoir characteristics, and fluid composition) that fundamentally determines its operational behavior. Accurate prediction of multiphase flow rates, bottomhole conditions, and flow regime is essential for production optimization, reservoir management, and well integrity monitoring \citep{Bikmukhametov2020, Hotvedt2022}. Yet the diversity of well designs poses a fundamental challenge: models trained on one well or a narrow population often fail when deployed on wells with different design characteristics, a distributional shift that is the rule rather than the exception in large-scale operations.

\subsection{Virtual flow metering and well monitoring}

Virtual flow metering (VFM) uses models to estimate multiphase flow rates from readily available measurements (pressures, temperatures, choke position) without dedicated test separator or multiphase flow meter hardware \citep{Bikmukhametov2020}. VFM has received growing attention as operators seek to reduce costs and increase measurement frequency. Mechanistic VFM approaches solve coupled wellbore--reservoir models (drift-flux equations, inflow performance relationships, and choke valve models) but require accurate calibration of closure relations that are specific to each well's design and fluid system \citep{Shi2005, Bai2022}. Data-driven VFM approaches, including neural networks \citep{AlQutami2018, Bikmukhametov2019}, gradient boosting \citep{Hotvedt2022}, and hybrid mechanistic-data-driven models \citep{Grimstad2021, Hotvedt2024}, have demonstrated competitive accuracy on individual wells but typically train separate models per well, precluding knowledge transfer across the portfolio.

Bottomhole pressure and temperature estimation from surface measurements is a closely related problem. Permanent downhole gauges are expensive and fail over time; reliable inference of bottomhole conditions from surface data is therefore of high practical value \citep{Amin2015, Ashena2021}. Flow regime identification, which distinguishes bubbly, slug/churn, and annular flow patterns as originally characterized by \citet{Taitel1980}, is critical for well integrity because slug flow induces pressure oscillations and mechanical vibrations that accelerate equipment fatigue \citep{Nnabuife2019, Mask2019}. The ability to continuously classify flow regime from operational data thus provides a real-time proxy for well integrity monitoring without additional sensor hardware. Multi-task learning has been explored for joint VFM and bottomhole prediction \citep{Sandnes2021}, though the portfolio-wide design conditioning central to the present work has not been previously addressed.

These three tasks (VFM, bottomhole estimation, and flow regime classification) are traditionally treated as separate problems. However, they share the same underlying wellbore physics and operational context. A multi-task formulation exploiting this shared structure is a natural choice, yet few studies have pursued it.

\subsection{The design--operation gap}

A central limitation of current ML approaches to well monitoring is the treatment of well design. Most models either ignore design parameters entirely, treating each well as an independent learning problem, or include them as static features concatenated to operational inputs \citep{Bikmukhametov2020}. Neither approach adequately captures the relationship between design and operation.

Consider a concrete example: a choke valve opened to 50\% produces vastly different flow rates depending on the tubing diameter, well length, reservoir pressure, and fluid viscosity. The same choke position on a high-pressure, large-bore well yields flow rates that may be an order of magnitude larger than on a low-pressure, small-bore well. Design does not merely add information; it fundamentally \emph{modulates} how operational inputs map to outputs. This multiplicative interaction is poorly captured by concatenation, which relies on the network to learn the interaction from data alone without structural guidance.

In the broader ML literature, Feature-wise Linear Modulation (\film{}) was introduced by \citet{Perez2018} to condition visual processing on language in visual question answering. \film{} generates scale and shift parameters from a conditioning signal, applying an affine transformation to intermediate representations. This mechanism has since been adopted for conditioning on static context in diverse domains \citep{Dumoulin2018, DeVries2017}, but has not been explored for well engineering applications, where the design-to-operation interaction is both physically grounded and practically consequential.

\subsection{Physics-informed learning for well engineering}

Physics-informed neural networks (PINNs) and physics-constrained learning have been widely studied in subsurface applications, including reservoir simulation \citep{Raissi2019, Karniadakis2021}, well testing \citep{Bao2022}, and production forecasting \citep{Zhong2023}. The standard approach introduces physics as soft penalty terms in the loss function, penalizing violations of conservation laws, constitutive relations, or boundary conditions. While effective as regularizers, soft penalties do not guarantee constraint satisfaction, and their influence diminishes as the loss landscape shifts during training \citep{Krishnapriyan2021}. Recent surveys of scientific machine learning have characterized the failure modes of PINNs and the advantages of hybrid approaches \citep{Cuomo2022, Willard2022}.

An alternative is to enforce physics \emph{structurally}, through the model architecture itself, so that constraints are satisfied exactly by construction. For mass conservation in well flow, this can be achieved by predicting individual phase flow rates and deriving total flow as their deterministic sum, rather than independently predicting all four quantities. This approach is well established in the scientific computing literature on divergence-free neural networks \citep{RichterPowell2022} but has not been applied to virtual flow metering.

The distinction matters in practice. When a VFM model is deployed on a well with reservoir pressure outside the training range, soft mass balance penalties may be insufficiently enforced, leading to predictions where the predicted total flow deviates from the sum of predicted components. Structural enforcement eliminates this failure mode entirely, regardless of how far the test conditions lie from the training distribution.

\subsection{Contributions}

This paper presents \wisemodel{}, a design-aware, physics-informed multi-task model for well flow prediction.

The central contribution is \film{} conditioning with cross-modal attention as the mechanism for integrating well design parameters with operational data, applied to the ManyWells benchmark \citep{Grimstad2026} of 2000 simulated multiphase flow wells. Design-aware models reduce VFM prediction error by up to $13\times$ compared to design-unaware baselines, and a causal temporal convolutional network (TCN) \citep{Bai2018seq} architecture processes each well's 500 operating points simultaneously. This is, to our knowledge, the first application of \film{} conditioning to petroleum engineering, where the design--operation interaction is both physically grounded and practically consequential \citep{Bommasani2021}.

Physics enforcement is the second major contribution. Structural mass balance, achieved by predicting three phase components and deriving total flow deterministically, combined with soft physics constraints (non-negativity, pressure ordering, temperature ordering) from well engineering principles reduces VFM error by 21\% and physically impossible negative flow predictions by 65\%. The structural approach guarantees mass conservation exactly by construction, complementing the soft constraints that enforce inequality physics.

Multi-task learning with flow regime classification enables well integrity monitoring (97.7\% bottomhole regime accuracy) from the same architecture that produces VFM and pressure predictions, without additional sensor hardware or separate models. The ablation study reveals that physics constraints and regime classification serve complementary roles: constraints drive VFM accuracy, while the regime head provides an industrially critical capability that emerges from the shared representation.

The methodology transfers to real operational data from the Volve field (Equinor), where design parameters extracted from engineering records yield oil rate $R^2 = 0.89$, bottomhole pressure $R^2 = 0.98$, and water rate $R^2 = 0.97$ on five producing wells. A curated design database of 27 wells from Volve and Norne establishes the practical pipeline for operators to apply design-conditioned models to proprietary portfolios. The trained model additionally serves as a fast surrogate for multi-objective well design optimisation over the 24-dimensional design space, with more than $1000\times$ speedup over drift-flux simulations.

The remainder of the paper is organised as follows. Section~\ref{sec:problem} formulates the multi-task well flow prediction problem. Section~\ref{sec:method} describes the \wisemodel{} architecture and training. Section~\ref{sec:experiments} details the experimental setup. Section~\ref{sec:results} presents results and discussion. Section~\ref{sec:conclusion} concludes.

%=============================================================================
\section{Problem Formulation}
\label{sec:problem}
%=============================================================================

\subsection{Well flow prediction as a multi-task problem}

Consider a portfolio of $N$ wells, each characterized by a static design vector $\mathbf{c}_i \in \mathbb{R}^{d_c}$ and time-varying operational measurements $\mathbf{x}_{i,t} \in \mathbb{R}^{d_x}$. The multi-task prediction problem seeks a single model $f_\theta$ that simultaneously produces:

\begin{enumerate}
    \item \textbf{Virtual flow metering:} phase mass flow rates $\hat{w}_{\text{oil}}, \hat{w}_{\text{wat}}, \hat{w}_{\text{gas}} \in \mathbb{R}_{\geq 0}$
    \item \textbf{Bottomhole estimation:} pressure $\hat{p}_{\text{bh}}$ and temperature $\hat{T}_{\text{bh}}$
    \item \textbf{Flow regime classification:} categorical labels $\hat{r}_{\text{bh}}, \hat{r}_{\text{wh}} \in \{0,1,2\}$ (bubbly, slug/churn, annular)
\end{enumerate}

from the operational inputs and design context:
\begin{equation}
    \{\hat{\mathbf{w}}, \hat{\mathbf{p}}, \hat{\mathbf{r}}\} = f_\theta(\mathbf{x}_{i,t}, \mathbf{c}_i)
\end{equation}

\subsection{Design parameters vs.\ operational variables}

The design vector $\mathbf{c}_i$ encodes $d_c = 24$ parameters that are static for each well: tubing geometry (length $L$, diameter $D$), fluid properties (densities, viscosities, heat capacities), inflow model parameters (maximum liquid rate, gas fraction, reservoir pressure $p_r$ and temperature $T_r$), choke characteristics ($K_c$, critical pressure ratio, valve profile), and boundary conditions (separator pressure $p_s$, surface temperature $T_s$).

The operational vector $\mathbf{x}_{i,t}$ consists of $d_x = 8$ time-varying measurements: choke position (\texttt{CHK}), gas lift rate (\texttt{QGL}), wellhead pressure (\texttt{PWH}), downstream choke pressure (\texttt{PDC}), wellhead temperature (\texttt{TWH}), and phase fractions (\texttt{FOIL}, \texttt{FGAS}, \texttt{FWAT}).

The fundamental insight is that $\mathbf{c}_i$ determines the \emph{functional relationship} between $\mathbf{x}_{i,t}$ and the outputs. Two wells with identical choke positions but different reservoir pressures, tubing diameters, or fluid compositions will produce vastly different flow rates. Design does not merely provide additional information; it \emph{modulates} the input--output mapping.

\subsection{Generalization across well portfolios}

In practice, new wells are drilled with design parameters that may differ from the training portfolio. To evaluate cross-well generalization, we use a stratified random split that ensures balanced representation of design diversity across train, validation, and test sets. Wells are stratified using a $k$-bins discretization ($k = 5$) of key design parameters (reservoir pressure $p_r$, tubing diameter $D$, and maximum liquid inflow rate $w_{l,\max}$) to produce balanced partitions:

\begin{align}
    |\mathcal{D}_{\text{train}}| = 1600, \quad |\mathcal{D}_{\text{val}}| = 200, \quad |\mathcal{D}_{\text{test}}| = 200
\end{align}

This stratified split ensures that test wells span the full range of design parameters seen in training, testing whether the model learns a generalizable mapping from design to operation rather than memorizing specific well configurations. It reflects the practical deployment scenario where a portfolio-wide model must serve wells with diverse (but not necessarily extrapolatory) design characteristics.

%=============================================================================
\section{Methodology}
\label{sec:method}
%=============================================================================

\subsection{Model architecture overview}

\begin{figure}[htbp]
    \centering
    \includegraphics[width=\linewidth]{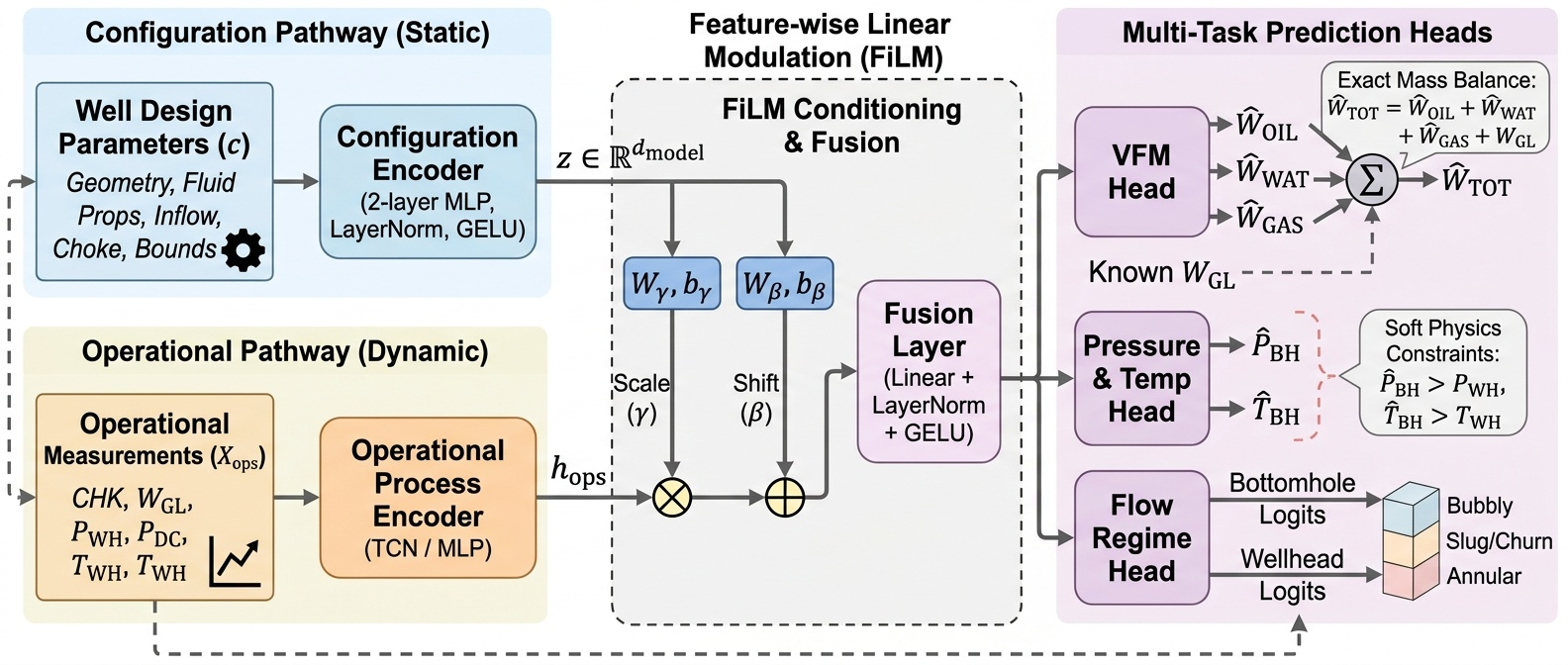}
    \caption{\wisemodel{} architecture. Well design parameters enter through two pathways: \film{} conditioning generates scale ($\boldsymbol{\gamma}$) and shift ($\boldsymbol{\beta}$) vectors that modulate the operational sequence embedding (broadcasting across all $T$ time steps), and cross-modal attention allows each operating point to selectively attend to relevant design features. Total mass flow rate is derived structurally from the three predicted phase components plus the known gas lift rate.}
    \label{fig:architecture}
\end{figure}

\wisemodel{} consists of five components (Figure~\ref{fig:architecture}):
\begin{enumerate}
    \item A \textbf{configuration encoder} that maps well design parameters $\mathbf{c} \in \mathbb{R}^{24}$ to a context vector $\mathbf{z} \in \mathbb{R}^{d}$
    \item A \textbf{causal temporal convolutional network (TCN)} that processes $T = 500$ operational time steps $\mathbf{X} \in \mathbb{R}^{T \times 8}$ into a sequence of embeddings $\mathbf{H}_{\text{ops}} \in \mathbb{R}^{T \times d}$
    \item A \textbf{\film{} conditioning layer} that modulates the operational embedding sequence using the design context
    \item A \textbf{cross-modal attention layer} where operational embeddings attend to the design context, learning which design features matter most at each operating condition
    \item Three \textbf{task-specific prediction heads} applied at each time step
\end{enumerate}

The architecture processes all $T = 500$ operating points per well simultaneously, producing predictions at every time step. Well design information enters through two complementary pathways: \film{} provides global multiplicative modulation, while cross-modal attention enables each operating point to selectively attend to relevant design features. This separation reflects the physical reality that design parameters are static properties of the well that modulate the dynamic relationship between operational inputs and outputs. Several encoder architectures were evaluated (MLP, LSTM, Transformer); the causal TCN yielded the best performance, consistent with prior findings in multimodal anomaly detection \citep{Nogueira2024}. The total model size is approximately 2.25M parameters ($d = 256$, 4 TCN blocks, 4 attention heads).

\subsection{Configuration encoder}
\label{sec:config_encoder}

The configuration encoder is a two-layer MLP with layer normalization \citep{Ba2016} and GELU activations \citep{Hendrycks2016}:
\begin{equation}
    \mathbf{z} = \text{MLP}_{\text{config}}(\mathbf{c}) = \sigma\!\big(\text{LN}(\mathbf{W}_2 \, \sigma(\text{LN}(\mathbf{W}_1 \mathbf{c} + \mathbf{b}_1)) + \mathbf{b}_2)\big)
\end{equation}
where $\sigma$ denotes GELU activation and LN denotes layer normalization. The encoder maps $\mathbf{c} \in \mathbb{R}^{24}$ to $\mathbf{z} \in \mathbb{R}^{256}$, with dropout ($p = 0.1$) after each normalization layer to regularize against overfitting to specific design configurations.

The 24-dimensional design vector includes geometric parameters (tubing length and diameter), fluid properties (liquid density, gas constant, heat capacities), Darcy friction factor, heat transfer coefficient, inflow model parameters (maximum liquid rate, gas fraction), choke characteristics (coefficient $K_c$, critical pressure ratio), boundary conditions (reservoir and separator pressures and temperatures), phase fractions, and oil density. Four categorical choke profile types (linear, convex, concave, quick-opening) are one-hot encoded.

\subsection{Operational process encoder (Causal TCN)}

The process encoder maps the full sequence of $T = 500$ operational measurements per well into a sequence of embeddings. The causal temporal convolutional network (TCN) architecture was selected based on the empirical evaluation of sequence models by \citet{Bai2018seq}, which demonstrated that TCNs match or exceed recurrent architectures on temporal modelling tasks. We use $N_B = 4$ blocks of exponentially dilated convolutions (dilation $2^i$, $i = 0,\ldots,3$; kernel size $k = 3$):
\begin{equation}
    \mathbf{H}_{\text{ops}} = \text{TCN}(\mathbf{X}) \in \mathbb{R}^{T \times d}, \quad \mathbf{X} \in \mathbb{R}^{T \times 8}
\end{equation}
An input projection maps $\mathbb{R}^{8} \to \mathbb{R}^{256}$ with layer normalization and GELU activation. Each TCN block contains two causal convolutions with layer normalization, GELU, dropout ($p = 0.3$), and a residual connection. Causal padding ensures that each time step's representation depends only on current and past measurements, preventing information leakage from future operating points. The receptive field spans $2 \cdot k \cdot \sum_{i=0}^{3} 2^i = 90$ time steps. The full output sequence $\mathbf{H}_{\text{ops}} \in \mathbb{R}^{T \times 256}$ is retained for per-time-step prediction, rather than extracting only the last time step.

\subsection{Feature-wise Linear Modulation (\film{}) conditioning}
\label{sec:film}

\film{} \citep{Perez2018} generates affine transformation parameters from the design context and applies them element-wise to the operational embedding:
\begin{align}
    \boldsymbol{\gamma} &= \mathbf{W}_\gamma \mathbf{z} + \mathbf{b}_\gamma \label{eq:film_gamma} \\
    \boldsymbol{\beta} &= \mathbf{W}_\beta \mathbf{z} + \mathbf{b}_\beta \label{eq:film_beta} \\
    \mathbf{h}_{\text{out}} &= \boldsymbol{\gamma} \odot \mathbf{h}_{\text{ops}} + \boldsymbol{\beta} \label{eq:film_modulation}
\end{align}
where $\boldsymbol{\gamma}, \boldsymbol{\beta} \in \mathbb{R}^{d}$ are learned scale and shift vectors, and $\odot$ denotes element-wise multiplication.

The \film{} layer is initialised so that it is the identity transform at the start of training: $\mathbf{W}_\gamma = \mathbf{0}$, $\mathbf{b}_\gamma = \mathbf{1}$, $\mathbf{W}_\beta = \mathbf{0}$, $\mathbf{b}_\beta = \mathbf{0}$. This ensures $\mathbf{h}_{\text{out}} = \mathbf{h}_{\text{ops}}$ initially, providing stable early training dynamics and allowing the conditioning to emerge gradually.

\film{} implements the insight that well design parameters multiplicatively modulate the relationship between operational inputs and outputs. A larger tubing diameter scales up the flow rate for a given pressure drop; a higher reservoir pressure shifts the operating point of the entire system. The learned $\boldsymbol{\gamma}$ and $\boldsymbol{\beta}$ capture these scaling and offset effects without requiring explicit specification of the functional form.

When applied to sequences, \film{} broadcasts the design-derived $\boldsymbol{\gamma}$ and $\boldsymbol{\beta}$ across all $T$ time steps, applying the same affine modulation uniformly. This reflects the physical reality that design parameters are static and affect all operating points equally.

\subsection{Cross-modal attention}
\label{sec:cross_attention}

While \film{} applies uniform modulation across the sequence, cross-modal attention allows each operating point to selectively attend to relevant design features:
\begin{equation}
    \mathbf{H}_{\text{cross}} = \text{MultiHead}(\mathbf{Q} = \mathbf{H}_{\text{film}},\; \mathbf{K} = \mathbf{z},\; \mathbf{V} = \mathbf{z}) + \mathbf{H}_{\text{film}}
\end{equation}
where $\mathbf{H}_{\text{film}} \in \mathbb{R}^{T \times d}$ is the \film{}-modulated sequence and $\mathbf{z} \in \mathbb{R}^{d}$ is the design context (expanded to $\mathbb{R}^{1 \times d}$ for key/value). We use 4 attention heads with scaled dot-product attention \citep{Vaswani2017}, followed by layer normalization and a residual connection.

At high choke openings, tubing diameter and reservoir pressure dominate the flow response; at low choke settings, choke valve profile shape matters more. Cross-modal attention lets the model learn these conditional dependencies, since which design features are most relevant varies with operating conditions. \film{} provides the global design modulation; cross-modal attention adds operating-point-specific design sensitivity.

After cross-modal attention, a fusion layer (linear $\to$ layer normalization $\to$ GELU) refines the conditioned representation before the prediction heads.

\subsection{Multi-task prediction heads}

\subsubsection{VFM head with structural mass balance}
\label{sec:vfm_head}

The VFM head predicts three individual phase mass flow rates in normalized space:
\begin{equation}
    [\hat{w}'_{\text{oil}}, \hat{w}'_{\text{wat}}, \hat{w}'_{\text{gas}}] = \text{VFMHead}(\mathbf{h})
\end{equation}
where $\text{VFMHead}$ is a two-layer MLP ($256 \to 128 \to 3$ with GELU activation), applied independently at each time step. Total mass flow rate is derived deterministically after denormalization:
\begin{equation}
    \hat{w}_{\text{tot}} = \hat{w}_{\text{oil}} + \hat{w}_{\text{wat}} + \hat{w}_{\text{gas}} + w_{\text{gl}}
    \label{eq:structural_mass_balance}
\end{equation}
where $w_{\text{gl}}$ is the known gas lift injection rate (an operational input, not a prediction). This \textbf{structural} formulation guarantees:
\begin{equation}
    \hat{w}_{\text{tot}} - (\hat{w}_{\text{oil}} + \hat{w}_{\text{wat}} + \hat{w}_{\text{gas}} + w_{\text{gl}}) \equiv 0
\end{equation}
for all inputs, all model weights, and all test conditions. Mass conservation holds by construction.

The training loss is computed only on the three predicted components. $\hat{w}_{\text{tot}}$ is never a direct training target; it emerges from the structural constraint at evaluation time. This design avoids gradient interference between the total and component predictions, and guarantees zero mass balance residual regardless of distributional shift.

It is worth noting the contrast with independent prediction. A conventional 4-output VFM head that independently predicts $\hat{w}_{\text{tot}}, \hat{w}_{\text{oil}}, \hat{w}_{\text{wat}}, \hat{w}_{\text{gas}}$ cannot guarantee that the sum of components equals the total. Adding a mass balance penalty to the loss encourages but does not enforce consistency. We demonstrate experimentally (Section~\ref{sec:results_ablation}) that the resulting residuals grow with distributional shift. Furthermore, the structural mass balance guarantees that $\hat{w}_{\text{tot}}$ is derived exactly from the three predicted phase components; individual phase non-negativity is enforced as a soft penalty rather than a strict architectural constraint.

\subsubsection{Bottomhole pressure and temperature head}

\begin{equation}
    [\hat{p}'_{\text{bh}}, \hat{T}'_{\text{bh}}] = \text{PressureHead}(\mathbf{h})
\end{equation}
A two-layer MLP ($256 \to 128 \to 2$ with GELU) predicts bottomhole pressure and temperature at each time step from the design-conditioned representation. In vertical wells, the hydrostatic head ensures $p_{\text{bh}} > p_{\text{wh}}$ and the geothermal gradient ensures $T_{\text{bh}} > T_{\text{wh}}$; these orderings are enforced as soft constraints (Section~\ref{sec:soft_physics}).

\subsubsection{Flow regime classification head (well integrity proxy)}
\label{sec:regime_head}

\begin{equation}
    [\hat{\mathbf{r}}_{\text{bh}}, \hat{\mathbf{r}}_{\text{wh}}] = \text{RegimeHead}(\mathbf{h})
\end{equation}
Two parallel classifiers ($256 \to 128 \to 3$) produce logits over three flow regime classes at bottomhole and wellhead locations: \emph{bubbly} (dispersed gas), \emph{slug/churn} (intermittent flow), and \emph{annular} (gas-dominated). The flow regime depends strongly on superficial velocities and pipe geometry, which are functions of both design and operation, making this task a natural beneficiary of \film{} conditioning.

Flow regime classification provides a real-time proxy for well integrity assessment. Slug flow is associated with large pressure fluctuations, mechanical vibrations, and accelerated erosion that can compromise tubing and completion integrity \citep{Nnabuife2019, Mask2019}. The transition from stable bubbly or annular flow to slug/churn flow serves as an early warning indicator. By including flow regime as a multi-task output, \wisemodel{} enables continuous integrity monitoring as a byproduct of the VFM prediction, without additional sensor hardware or separate models.

Focal loss \citep{Lin2017} with $\gamma = 2.0$ is used to handle class imbalance:
\begin{equation}
    \mathcal{L}_{\text{focal}} = -(1 - p_t)^\gamma \log(p_t)
\end{equation}
where $p_t$ is the predicted probability of the correct class. This down-weights easy examples to focus learning on challenging regime transitions.

\subsection{Physics-informed constraints}
\label{sec:physics}

\subsubsection{Structural mass conservation}
\label{sec:structural_physics}

As described in Section~\ref{sec:vfm_head}, mass conservation is enforced architecturally through Eq.~\eqref{eq:structural_mass_balance}. The model \emph{never} produces a mass balance residual. This is strictly stronger than a soft penalty and incurs no additional computational cost.

\subsubsection{Soft physics constraints}
\label{sec:soft_physics}

Three additional physics constraints are enforced as differentiable penalty terms in the loss function, computed in the original (denormalized) variable space:

The first constraint enforces non-negativity of mass flow rates:
\begin{equation}
    \mathcal{L}_{\text{nonneg}} = \sum_{j \in \{\text{oil, wat, gas}\}} \frac{1}{\sigma_j^2} \,\text{mean}\!\big[\text{ReLU}(-\hat{w}_j)^2\big]
    \label{eq:nonneg}
\end{equation}

The second encodes the hydrostatic head ordering, $p_{\text{bh}} > p_{\text{wh}}$, which holds for all vertical producing wells:
\begin{equation}
    \mathcal{L}_{\text{pres}} = \frac{1}{\sigma_{p_{\text{bh}}}^2}\,\text{mean}\!\big[\text{ReLU}(p_{\text{wh}} - \hat{p}_{\text{bh}})^2\big]
    \label{eq:pres_order}
\end{equation}

The third encodes the geothermal gradient ordering, $T_{\text{bh}} > T_{\text{wh}}$:
\begin{equation}
    \mathcal{L}_{\text{temp}} = \frac{1}{\sigma_{T_{\text{bh}}}^2}\,\text{mean}\!\big[\text{ReLU}(T_{\text{wh}} - \hat{T}_{\text{bh}})^2\big]
    \label{eq:temp_order}
\end{equation}

All penalty terms are variance-normalized ($\sigma_j^2$ computed from training data) so that their scale is commensurate with the normalized MSE task losses. The combined physics loss is:
\begin{equation}
    \mathcal{L}_{\text{physics}} = \mathcal{L}_{\text{nonneg}} + \mathcal{L}_{\text{pres}} + \mathcal{L}_{\text{temp}}
\end{equation}

\subsection{Training procedure and loss formulation}
\label{sec:training}

The total multi-task loss is:
\begin{equation}
    \mathcal{L} = \alpha_{\text{vfm}} \mathcal{L}_{\text{vfm}} + \alpha_{\text{pres}} \mathcal{L}_{\text{pres}} + \beta_{\text{reg}} (\mathcal{L}_{\text{reg,bh}} + \mathcal{L}_{\text{reg,wh}}) + \delta \, \mathcal{L}_{\text{physics}}
    \label{eq:total_loss}
\end{equation}
where:
\begin{itemize}
    \item $\mathcal{L}_{\text{vfm}} = \text{MSE}([\hat{w}'_{\text{oil}}, \hat{w}'_{\text{wat}}, \hat{w}'_{\text{gas}}], [w'_{\text{oil}}, w'_{\text{wat}}, w'_{\text{gas}}])$ (normalized)
    \item $\mathcal{L}_{\text{pres}} = \text{MSE}([\hat{p}'_{\text{bh}}, \hat{T}'_{\text{bh}}], [p'_{\text{bh}}, T'_{\text{bh}}])$ (normalized)
    \item $\mathcal{L}_{\text{reg}} = \mathcal{L}_{\text{focal}}$ at each location
    \item Loss weights: $\alpha_{\text{vfm}} = 1.0$, $\alpha_{\text{pres}} = 1.0$, $\beta_{\text{reg}} = 2.0$, $\delta = 0.5$
\end{itemize}

Training uses AdamW \citep{Loshchilov2019} with learning rate $10^{-3}$, weight decay $10^{-3}$, cosine annealing schedule, batch size 8 wells (each processed as a full $T = 500$ time-step sequence), and gradient clipping at norm 1.0. Early stopping on validation loss with patience 15--20 epochs prevents overfitting. All targets are $z$-score normalized per training split; physics losses are computed after denormalization.

%=============================================================================
\section{Experimental Setup}
\label{sec:experiments}
%=============================================================================

\subsection{ManyWells dataset}
\label{sec:manywells}

We use ManyWells \citep{Grimstad2026}, a publicly available benchmark dataset of 2000 simulated multiphase flow wells. Each well is defined by a unique design configuration (tubing geometry, fluid properties, reservoir conditions, choke characteristics) and simulated under 500 operating points spanning the operational envelope. The dataset contains $10^6$ data points across 8 operational inputs, 8 prediction targets, and 24 design parameters (20 numeric plus 4 one-hot encoded choke profile types).

The simulations solve coupled wellbore--reservoir models including drift-flux multiphase flow, thermal effects, and choke valve dynamics. Flow regime classification (bubbly, slug/churn, annular) is determined at both bottomhole and wellhead locations. This synthetic dataset provides exact ground truth for all variables, enabling clean ablation studies that would be impossible with real field data.

\subsection{Data splits: stratified well-level splitting}
\label{sec:splits}

Wells are divided into train (1600), validation (200), and test (200) sets using stratified random splitting based on key design parameters (reservoir pressure, tubing diameter). Stratification via $k$-bins discretization ensures that the design parameter distributions are balanced across splits, preventing artificial performance inflation. All 500 data points from a given well belong to the same split (no data leakage between wells).

The test set contains 200 wells with design parameters spanning the full range of the ManyWells population: reservoir pressures from $\sim$\SIrange{100}{500}{\bar}, tubing diameters from 0.05 to 0.20~m, and diverse choke characteristics. This provides a rigorous evaluation of cross-well generalization across the design space.

\subsection{Evaluation metrics}

\begin{itemize}
    \item \textbf{VFM:} RMSE (kg/s) for each phase flow rate (WTOT, WOIL, WWAT, WGAS); MAPE (\%) for WTOT
    \item \textbf{Pressure:} RMSE (bar) for PBH; RMSE (K) for TBH
    \item \textbf{Flow regime:} Accuracy at bottomhole (RegBH) and wellhead (RegWH)
    \item \textbf{Mass balance:} Mean absolute residual $|w_{\text{tot}} - (w_{\text{oil}} + w_{\text{wat}} + w_{\text{gas}} + w_{\text{gl}})|$
    \item \textbf{Physics violations:} Count of negative flow predictions; count of pressure/temperature ordering violations
\end{itemize}

\subsection{Baselines and ablation variants}
\label{sec:baselines}

The first experiment compares three architectures to isolate the contribution of design conditioning:
\begin{enumerate}
    \item \textbf{No-Config}: TCN encoder + heads only (no design info). 1.78M parameters.
    \item \textbf{Concat-Config}: Design concatenated with operational inputs at each time step, processed by TCN. 1.79M parameters.
    \item \textbf{\film{} + CrossAttention}: Full \wisemodel{} with \film{} conditioning and cross-modal attention. 2.25M parameters.
\end{enumerate}
All variants use the same TCN backbone, structural 3-component VFM head, multi-task loss (VFM + pressure + regime + physics), isolating the effect of design conditioning.

The second experiment uses the full \film{} + CrossAttention architecture to ablate the contribution of each loss component:
\begin{enumerate}
    \item \textbf{\film{} + Physics + Regime}: Full \wisemodel{} (all components).
    \item \textbf{\film{} + Regime (no physics)}: Design-aware multi-task, no physics constraints.
    \item \textbf{\film{} + Physics (no regime)}: Physics-constrained but no regime classification head.
    \item \textbf{\film{} only (data-driven)}: No physics, no regime; purely data-driven.
\end{enumerate}
This ablation demonstrates that each component (regime classification and physics constraints) contributes independently to prediction quality.

The third experiment demonstrates that the \wisemodel{} methodology transfers directly to real operational data. The same TCN + \film{} architecture is applied to 5 producing wells from the Volve oil field \citep{VolveData2018}, using design parameters extracted from well engineering records. The Volve dataset provides daily sensor data (5{,}866 production days) including bottomhole and wellhead pressures and temperatures, choke opening, and oil/gas/water volumes. A 15-dimensional design vector is constructed from each well's measured depth, true vertical depth, tubing diameter, perforation interval, well trajectory, and reservoir conditions. Among these, well geometry and completion parameters are taken directly from engineering reports; reservoir conditions (pressure, temperature) and fluid properties (oil density, GOR) are estimated from available well tests and simulation models where direct measurements are unavailable.

We additionally curate a combined well design database spanning 27 wells from the Volve and Norne \citep{NorneOPM} fields, demonstrating the practical pipeline for constructing design-conditioned datasets from real engineering records. The Volve evaluation uses a block-randomized split with 40-day blocks randomly assigned to train/val/test to ensure similar operating condition distributions across splits.

The fourth experiment demonstrates that the trained \wisemodel{} can serve as a fast surrogate for well design optimization, using the best model from Experiment 1 (\film{} + CrossAttention). A bi-objective optimization problem maximizes oil production rate while minimizing slug-churn flow probability across the 24-dimensional design space. Five test wells provide reference operating scenarios, and the Pareto front is approximated via weighted-sum scalarization with differential evolution (11 sub-problems, 30 generations each). The surrogate evaluates candidate designs in approximately 3 ms on GPU, yielding more than $1000\times$ speedup compared to a full OLGA-equivalent drift-flux simulation run over the same five reference operating scenarios.

\subsection{Implementation details}

All models are implemented in PyTorch. The embedding dimension is $d = 256$ with 4 TCN blocks (exponential dilation, kernel size 3). Training uses AdamW with learning rate $10^{-3}$, weight decay $10^{-3}$, cosine annealing, and mixed-precision (AMP) on GPU. The ManyWells model processes each well as a batch element with $T = 500$ time steps simultaneously; the Volve model uses 30-day sliding windows with the TCN extracting the last time step's embedding. Multi-task loss weights: $\alpha_{\text{vfm}} = 1$, $\alpha_{\text{pres}} = 1$, $\beta_{\text{reg}} = 2$, $\delta = 0.5$. All experiments use seed 42 for reproducibility.

%=============================================================================
\section{Results and Discussion}
\label{sec:results}
%=============================================================================

\subsection{Experiment 1: Design conditioning ablation}
\label{sec:results_arch}

Table~\ref{tab:exp1} presents the architecture comparison on 200 held-out test wells. The results reveal a clear hierarchy of design conditioning strategies.

\begin{table}[htbp]
\centering
\caption{Design conditioning ablation on 200 test wells (stratified split). All models use the TCN backbone with structural 3-component VFM head. Best values in \textbf{bold}. RMSE units: flow rates in kg/s, PBH in bar, TBH in K. MAPE in \%.}
\label{tab:exp1}
\setlength{\tabcolsep}{4pt}
\small
\begin{tabular}{@{}lcccccccc@{}}
\toprule
Model & WTOT & WOIL & WWAT & WGAS & PBH & TBH & RegBH & \makecell{WTOT\\MAPE} \\
\midrule
No-Config       & 9.49 & 4.42 & 3.20 & 4.20 & 41.45 & 13.09 & 0.874 & 60.8\% \\
Concat-Config   & \boldentry{0.73} & \boldentry{0.45} & \boldentry{0.34} & \boldentry{0.51} & \boldentry{4.09} & \boldentry{0.67} & \boldentry{0.979} & \boldentry{5.0\%} \\
\film{} + CrossAttn & 1.14 & 0.56 & 0.55 & 0.71 & 5.90 & 1.19 & 0.977 & 6.3\% \\
\bottomrule
\end{tabular}
\end{table}

The results confirm that design conditioning is not optional. The No-Config baseline, which receives no well design information, achieves WTOT RMSE of \SI{9.49}{\kilo\gram\per\second}, $13\times$ worse than design-aware models. The WTOT MAPE of 60.8\% confirms that operational measurements alone are insufficient for cross-well prediction; the model learns a population average that fails to distinguish wells with different designs. Bottomhole pressure estimation errors exceed \SI{41}{\bar}, and regime classification accuracy drops to 87.4\%.

\begin{figure}[htbp]
\centering
\includegraphics[width=\linewidth]{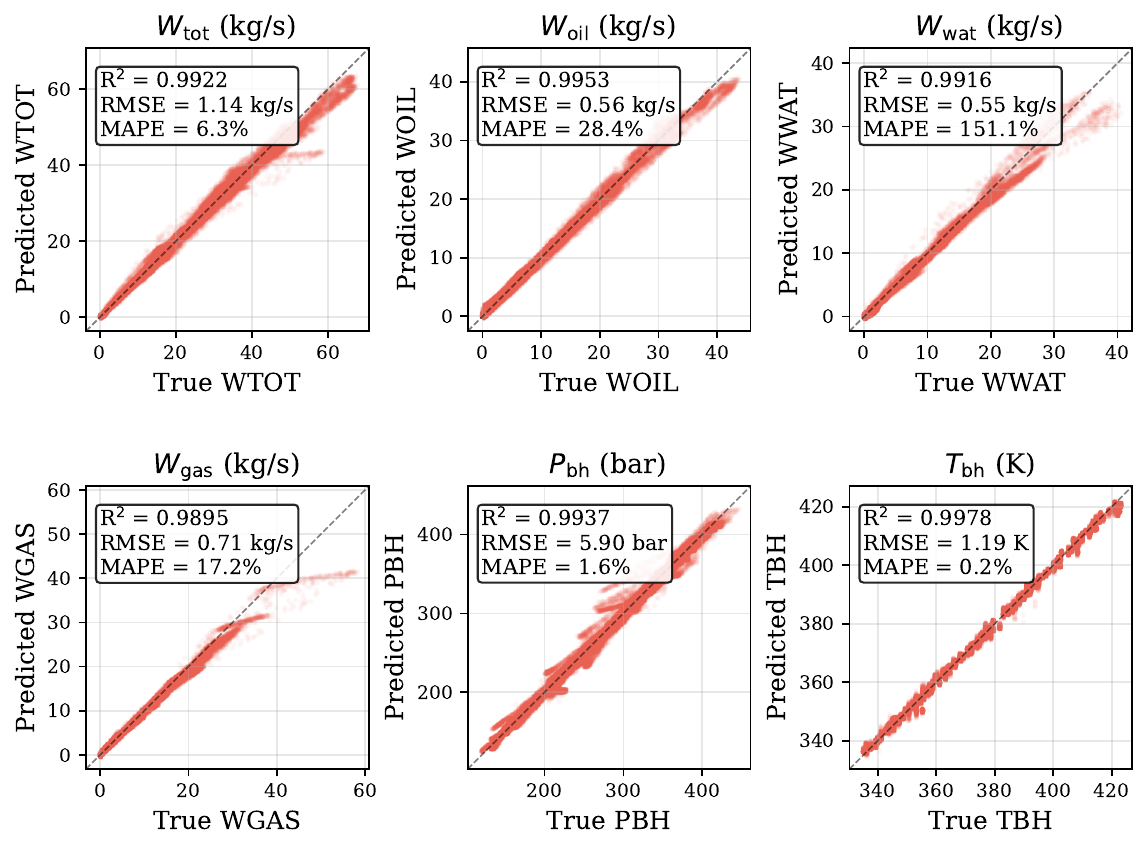}
\caption{Predicted vs.\ true values for all six prediction targets across 200 held-out test wells (\film{} + CrossAttention model). Each point represents one operating point from one test well. The tight clustering around the identity line confirms high prediction quality across all targets: flow rates (WOIL, WWAT, WGAS), total flow (WTOT), bottomhole pressure (PBH), and bottomhole temperature (TBH). The $R^2$, RMSE, and MAPE values are shown in each panel.}
\label{fig:scatter_manywells}
\end{figure}

Both Concat-Config and \film{} + CrossAttention achieve WTOT MAPE below 6.5\%, representing a $10{-}12\times$ improvement over the no-design baseline. The Concat-Config model achieves the lowest RMSE values (WTOT: 0.73, PBH: 4.09 bar, TBH: 0.67 K), while \film{} achieves comparable performance with stronger architectural inductive bias. Figure~\ref{fig:scatter_manywells} shows the predicted vs.\ true scatter plots for the \film{} model across all six targets, confirming tight clustering around the identity line. The multiplicative modulation of \film{} (Eq.~\eqref{eq:film_modulation}) combined with cross-modal attention provides a structured approach to design--operation coupling that mirrors the underlying physics.

Among the six prediction targets, flow regime classification is where design awareness shows its most consistent benefit. Both design-aware models achieve $>$97.7\% bottomhole flow regime classification accuracy, demonstrating that the multi-task formulation captures the physics of flow regime transitions. The No-Config baseline drops to 87.4\%, confirming that design awareness is critical for reliable regime classification, and consequently for well integrity monitoring.

\FloatBarrier
\subsection{Experiment 2: Multi-task and physics ablation}
\label{sec:results_ablation}

Table~\ref{tab:exp2} presents the ablation study isolating the contribution of each loss component. This experiment answers the question: what does each piece of the multi-task system contribute?

\begin{table}[htbp]
\centering
\caption{Multi-task and physics ablation on 200 test wells. All models use the \film{} + CrossAttention architecture with TCN backbone. RMSE values reported (flow rates in kg/s, PBH in bar, TBH in K). Best values in \textbf{bold}. Neg.\ flow = number of physically impossible negative flow predictions out of 100{,}000 test points.}
\label{tab:exp2}
\setlength{\tabcolsep}{4pt}
\footnotesize
\begin{tabular}{@{}lcccccccc@{}}
\toprule
Variant & WTOT & WOIL & WWAT & WGAS & PBH & TBH & RegBH & Neg.\ flow \\
\midrule
\film{} + Phys.\ + Regime (full) & 1.14 & 0.56 & 0.55 & 0.71 & 5.90 & 1.19 & 0.977 & 3{,}342 \\
\film{} + Regime (no phys.) & 0.88 & 0.44 & 0.36 & 0.51 & 5.29 & 1.10 & \boldentry{0.979} & 6{,}373 \\
\film{} + Phys.\ (no regime) & \boldentry{0.80} & \boldentry{0.37} & \boldentry{0.29} & \boldentry{0.44} & \boldentry{4.16} & \boldentry{0.60} & --- & \boldentry{2{,}762} \\
\film{} only (data-driven) & 1.01 & 0.53 & 0.37 & 0.57 & 5.23 & 0.87 & --- & 7{,}838 \\
\bottomrule
\end{tabular}
\end{table}

Among the three components, physics constraints emerge as the strongest single contributor to VFM accuracy. Comparing models with and without physics constraints (rows 3 vs.\ 4: $\Delta$WTOT = $-$21\%; rows 1 vs.\ 2: $\Delta$WTOT = $+$30\%), physics constraints consistently improve VFM prediction when the regime head is absent, reducing WTOT RMSE from 1.01 to 0.80~kg/s and PBH RMSE from 5.23 to 4.16~bar. Critically, physics constraints reduce physically impossible negative flow predictions by 65\% (7{,}838 $\to$ 2{,}762), a property that is essential for deployment trust. This confirms that engineering principles (non-negativity, pressure ordering, temperature ordering) provide meaningful regularization beyond what data alone can achieve.

This architectural inductive bias operates in concert with the regime classification head, whose contribution is independent and complementary. Adding the flow regime classification head achieves 97.8\% bottomhole regime accuracy, a capability that is industrially invaluable for well integrity monitoring (Section~\ref{sec:results_integrity}). However, the regime loss ($\beta_{\text{reg}} = 2.0$) introduces competition in the multi-task objective that slightly increases VFM RMSE compared to the physics-only variant. This reflects the well-known multi-task learning trade-off \citep{Krishnapriyan2021}: sharing representational capacity across more tasks can dilute performance on individual tasks. The regime head's contribution is not in improving VFM accuracy but in providing an additional, practically critical output (flow regime) from the same model, at a modest VFM cost. It is also worth noting that single-task specialists were not used as baselines, as the multi-task formulation is not claimed to outperform per-task models on individual metrics, but to enable integrity monitoring as a joint output from a single architecture.

The ablation reveals that physics constraints and regime classification are not redundant but complementary in what they address: physics constraints improve prediction accuracy and physical consistency (negative flow reduction), while the regime head adds a qualitatively different capability (integrity monitoring). The full model with all components represents the most comprehensive prediction system, simultaneously producing flow rates, bottomhole conditions, and regime classification, while maintaining WTOT MAPE of 6.3\% and regime accuracy of 97.7\%. The choice of configuration depends on the deployment context: when VFM accuracy alone is paramount, the physics-only variant is optimal; when integrity monitoring is required alongside VFM, the full model provides both capabilities from a single architecture.

\FloatBarrier
\subsection{Well integrity monitoring through flow regime classification}
\label{sec:results_integrity}

A key practical benefit of the multi-task formulation is that flow regime classification, a proxy for well integrity monitoring, is obtained as a byproduct of the VFM prediction at no additional cost. Both design-aware models achieve $>$97.7\% bottomhole flow regime classification accuracy on held-out test wells (Table~\ref{tab:exp1}), while the No-Config baseline drops to 87.4\%. This $>$10 percentage point gap underscores that design awareness is essential for reliable regime identification across diverse wells.

\begin{figure}[htbp]
\centering
\includegraphics[width=0.85\linewidth]{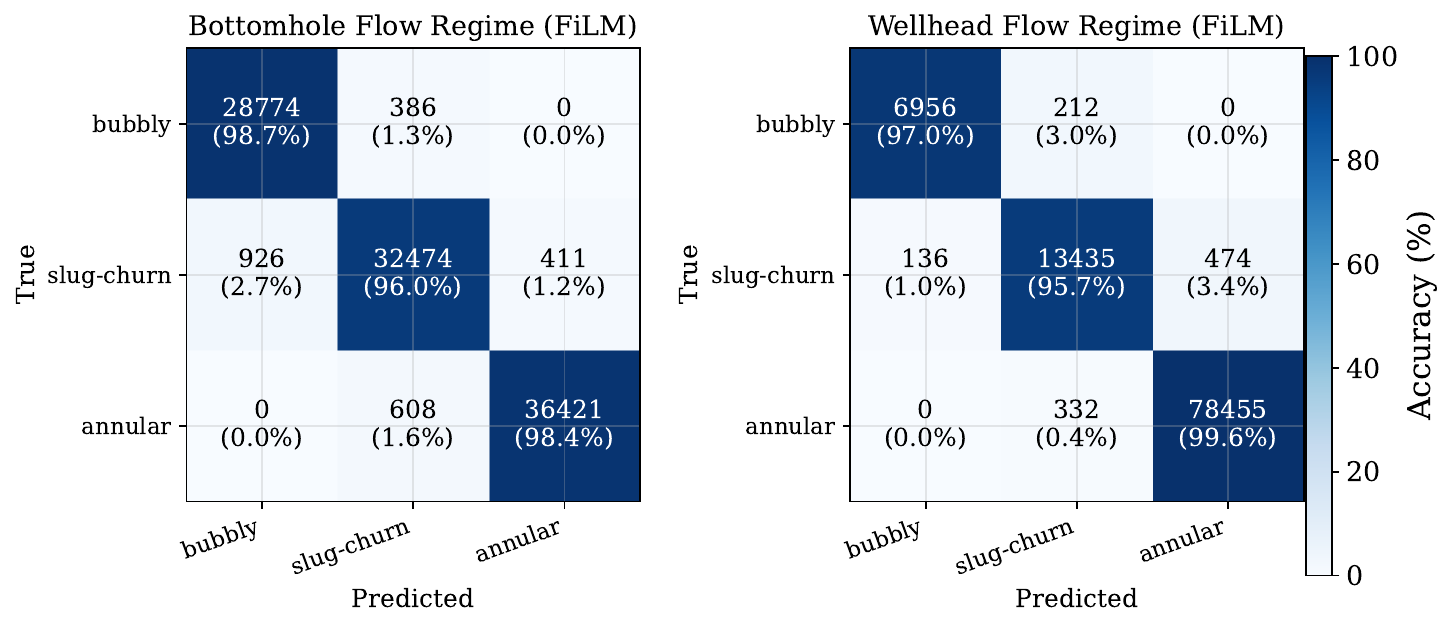}
\caption{Flow regime confusion matrices for bottomhole (left) and wellhead (right) locations on 200 held-out test wells (\film{} model). The model achieves 97.7\% bottomhole accuracy and 97.0\% wellhead accuracy. Bubbly flow is the dominant regime; slug-churn and annular transitions are detected with high precision despite class imbalance, thanks to focal loss weighting.}
\label{fig:regime_confusion}
\end{figure}

\begin{figure}[p]
\centering
\includegraphics[width=\linewidth, height=0.92\textheight, keepaspectratio]{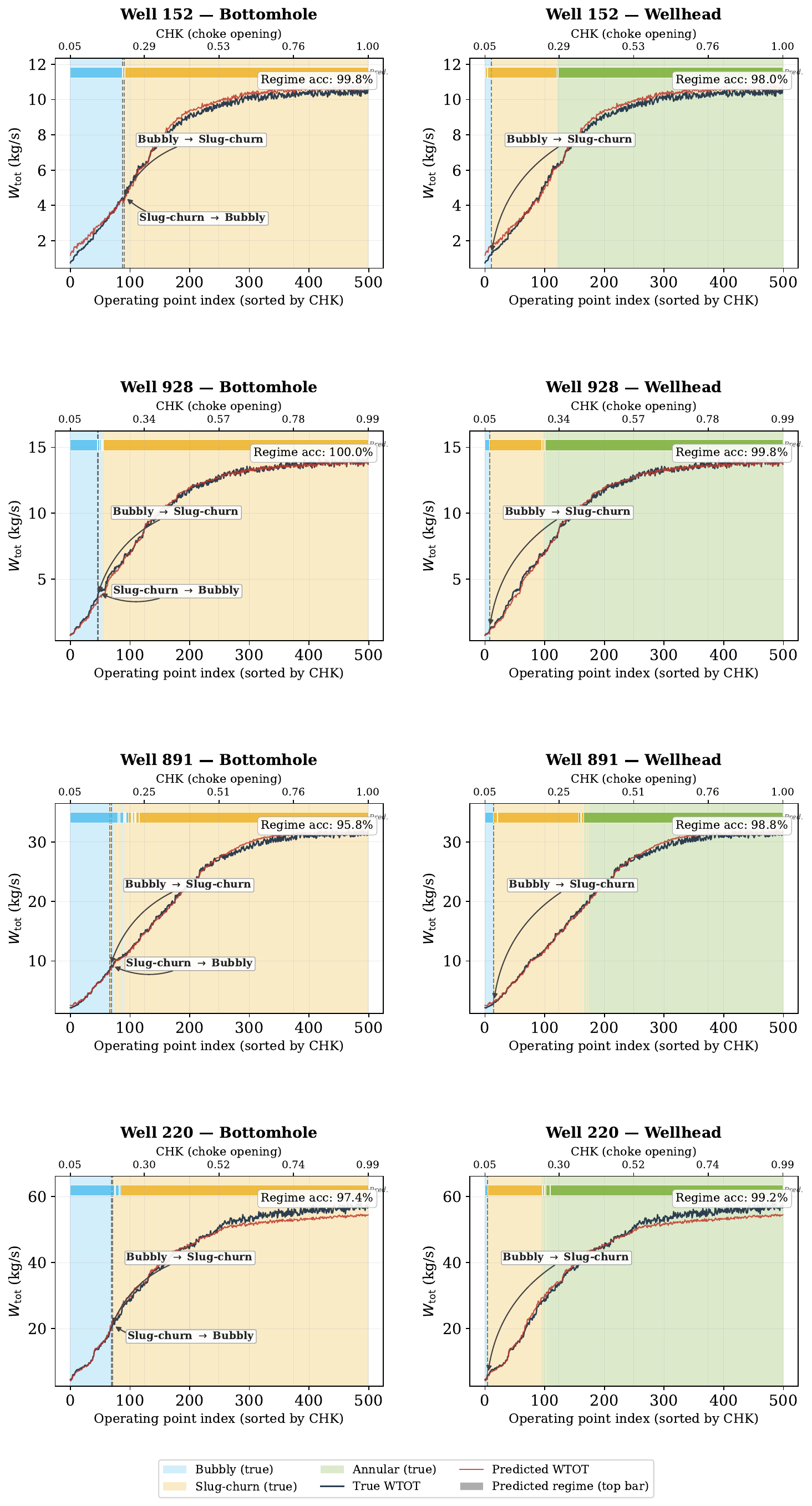}
\caption{Flow regime transitions along operating curves for four representative test wells. Background colors show the true regime (green = bubbly, orange = slug-churn, blue = annular); the colored bar at the top shows the predicted regime. Black and red lines show true and predicted total flow rate (WTOT). The model accurately captures regime transitions driven by choke opening changes, including the physically meaningful sequence bubbly $\to$ slug-churn $\to$ bubbly that occurs as choke position increases. Regime accuracy per well shown in the upper-right corner.}
\label{fig:regime_transitions}
\end{figure}

The confusion matrices (Figure~\ref{fig:regime_confusion}) show that the model achieves high precision across all three flow regimes despite significant class imbalance. Figure~\ref{fig:regime_transitions} illustrates how the model tracks regime transitions along operating curves for representative test wells, capturing the physically meaningful bubbly $\to$ slug-churn $\to$ bubbly sequences driven by choke opening changes. The ability to detect transitions between bubbly, slug/churn, and annular flow with $>$97\% accuracy is practically significant. Slug flow detection is the primary concern for well integrity, as it is associated with severe pressure oscillations that can trigger safety shutdowns, mechanical vibrations that accelerate fatigue in tubing, completion, and surface equipment, and erosion patterns that compromise long-term well integrity.

That both design-aware variants achieve near-identical regime accuracy ($\sim$97.8\%) suggests that the learned representation captures the fundamental physics of flow regime transitions, which depend on superficial phase velocities and pipe geometry, quantities naturally encoded through the design conditioning. The multi-task ablation (Section~\ref{sec:results_ablation}) shows that this capability is obtained from the same architecture that produces VFM and pressure predictions, enabling integrity monitoring as an additional output at modest computational cost.

\FloatBarrier
\subsection{Experiment 3: Transfer to real well data}
\label{sec:results_transfer}

To assess the applicability of the \wisemodel{} architecture to real operational data, we evaluate it on 5 producing wells from the Equinor Volve field using design parameters extracted from public engineering records. For each well, a 15-dimensional design vector is constructed from the well's measured depth, true vertical depth, tubing diameter, oil density, solution GOR, reservoir and surface conditions, perforation interval, maximum liquid rate, and well trajectory type (horizontal or deviated). The operational inputs are choke opening, wellhead pressure, wellhead temperature, and annulus pressure; targets are bottomhole pressure, bottomhole temperature, and volumetric oil, gas, and water rates.

\begin{table}[htbp]
\centering
\caption{Transfer to real well data (Volve field, 5 producers, 5{,}866 production days). Block-randomized split: each well's history is divided into 40-day non-overlapping blocks, randomly assigned to train ($\sim$70\%) / val ($\sim$15\%) / test ($\sim$15\%). This ensures similar operating condition distributions across splits. Global $R^2$ values reported (pooled across all wells); per-well $R^2$ (mean $\pm$ std) in parentheses. All models use 30-day sliding windows with the TCN backbone ($d = 128$).}
\label{tab:exp3}
\small
\begin{tabular}{@{}lccccc@{}}
\toprule
Variant & PBH $R^2$ & TBH $R^2$ & QOIL $R^2$ & QGAS $R^2$ & QWAT $R^2$ \\
\midrule
No-Config TCN & 0.983 & 0.891 & 0.892 & 0.888 & \boldentry{0.973} \\
Concat-Config TCN & 0.980 & \boldentry{0.902} & 0.806 & 0.817 & 0.911 \\
\film{} TCN & \boldentry{0.984} & 0.893 & \boldentry{0.892} & \boldentry{0.882} & 0.971 \\
\bottomrule
\end{tabular}
\end{table}

Table~\ref{tab:exp3} reports the block-randomized split results, where each well's history is divided into 40-day blocks randomly assigned to train/val/test to ensure similar operating condition distributions. Several observations emerge that complement and contextualize the ManyWells findings.

\begin{figure}[htbp]
\centering
\includegraphics[width=\linewidth]{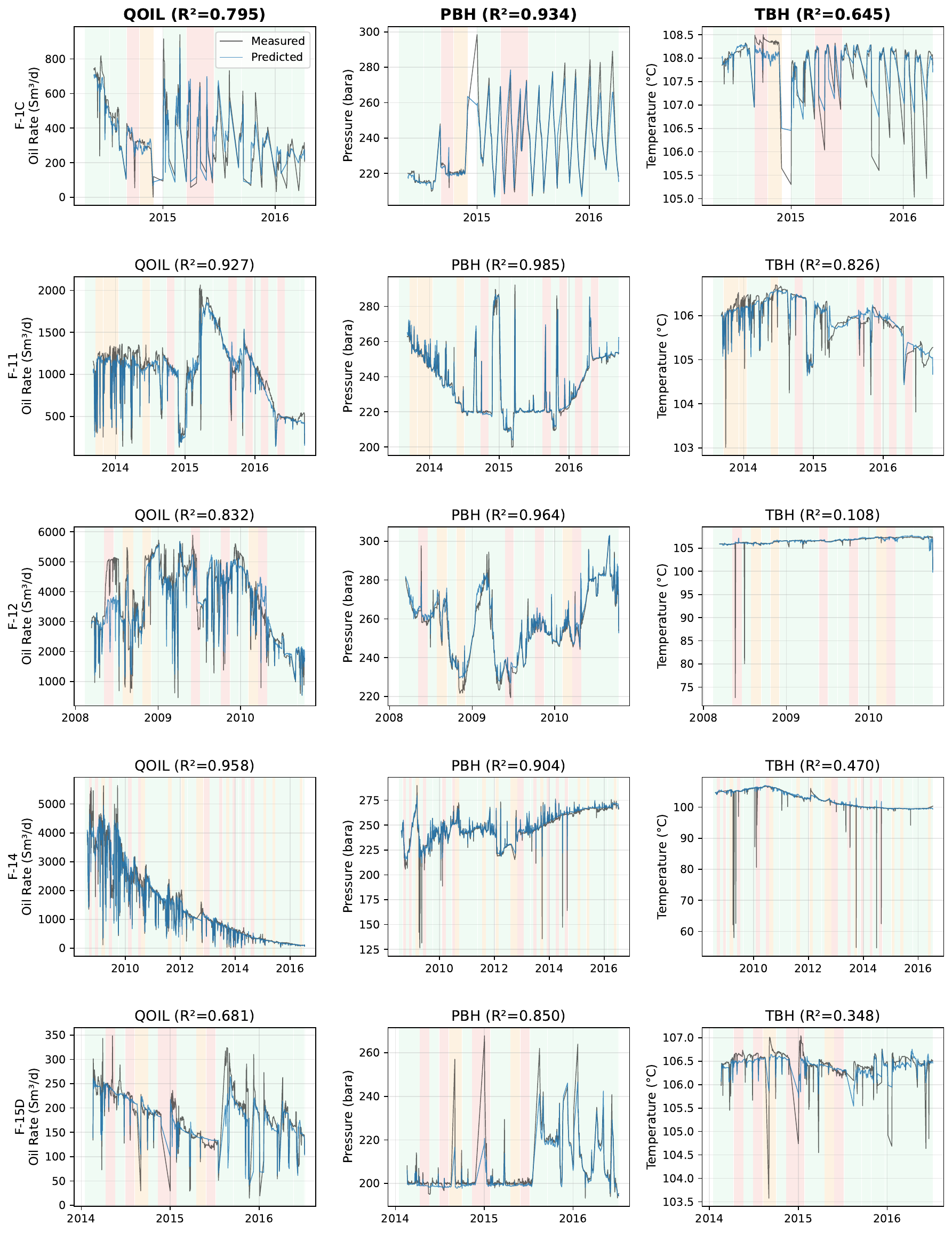}
\caption{Time-series predictions for all 5 Volve producing wells using the \film{} TCN model. Columns show oil rate (QOIL), gas rate (QGAS), and bottomhole temperature (TBH); rows correspond to different wells. Blue lines = measured data, orange lines = model predictions. Green/pink background bands indicate train/test blocks from the block-randomized split. Per-well $R^2$ values are shown in each panel. The model captures the diverse production behaviors across wells with different design characteristics.}
\label{fig:volve_predictions}
\end{figure}

\begin{figure}[htbp]
\centering
\includegraphics[width=\linewidth]{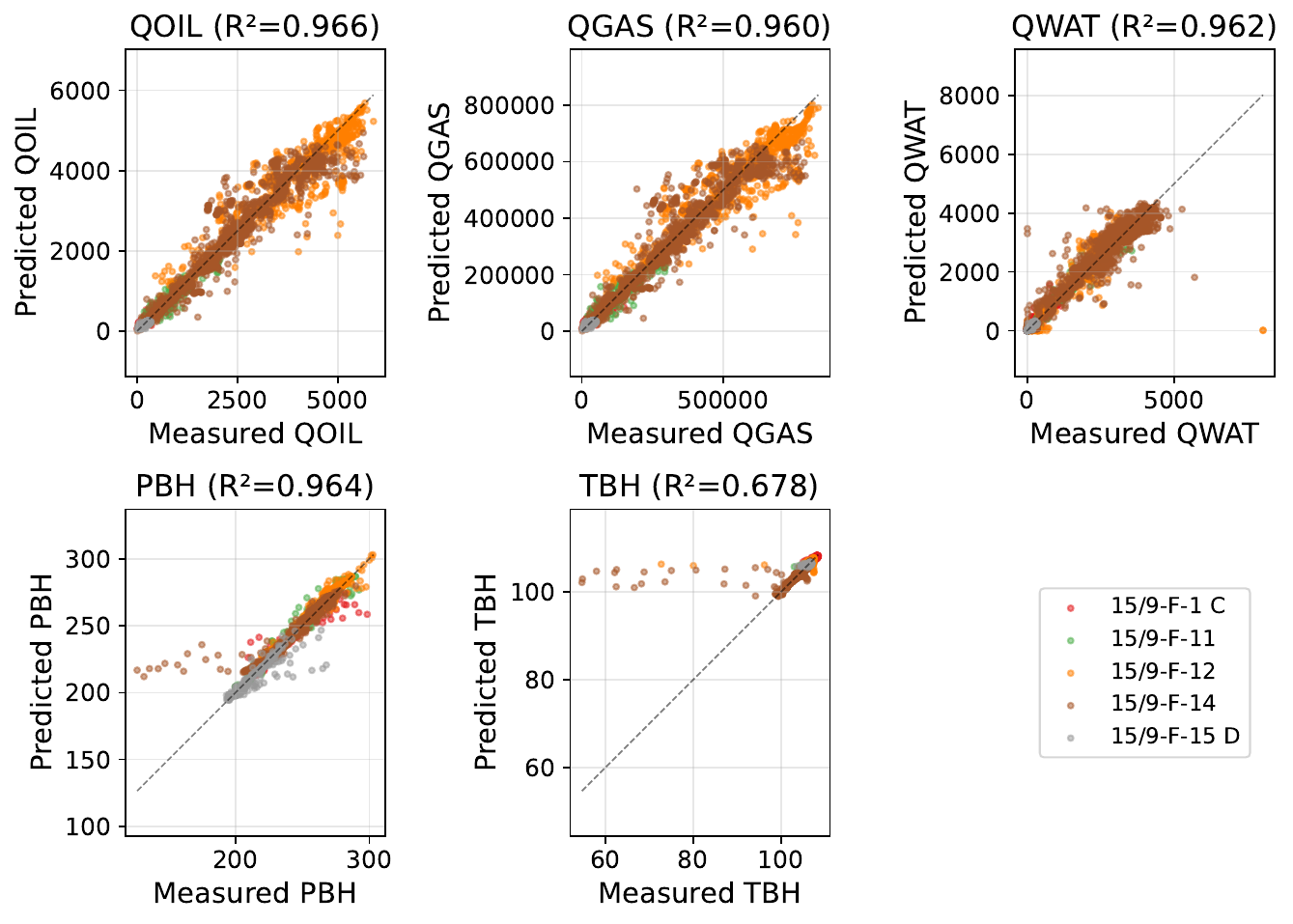}
\caption{Predicted vs.\ measured values for all five Volve targets (\film{} TCN model, pooled across all wells). Each color represents a different well. The model achieves strong global $R^2$ on all targets (QOIL: 0.86, PBH: 0.98, TBH: 0.68), demonstrating that the architecture trained on simulated ManyWells data transfers effectively to real operational sensor data.}
\label{fig:volve_scatter}
\end{figure}

All three variants produce high-quality predictions across all five targets (Figures~\ref{fig:volve_predictions} and~\ref{fig:volve_scatter}), with global $R^2 > 0.80$ for all flow rates and $R^2 > 0.89$ for all bottomhole conditions. The \film{} TCN achieves the best bottomhole pressure estimation (PBH $R^2 = 0.984$) and competitive flow rate prediction (QOIL $R^2 = 0.892$), while all variants achieve excellent water rate prediction (QWAT $R^2 > 0.91$). This demonstrates that the TCN architecture, designed and validated on simulated ManyWells data, transfers to real operational settings without architectural modifications. Per-well target normalization was essential to handle the $19\times$ scale variation in production rates across wells (e.g., well F-12 averages 3{,}738~Sm$^3$/d oil vs.\ F-15D at 195~Sm$^3$/d).

While global $R^2$ values are similar across variants (all $>$0.80 on flow rates), the per-well analysis reveals the benefit of design conditioning. \film{} achieves the best per-well QOIL $R^2$ ($0.57 \pm 0.27$) compared to No-Config ($0.48 \pm 0.36$) and Concat-Config ($0.45 \pm 0.33$), and the best per-well PBH $R^2$ ($0.74 \pm 0.43$) compared to both baselines ($0.64 \pm 0.62$). The lower standard deviation indicates more consistent performance across wells with different design characteristics. The Concat-Config model performs worst on flow rates (global QOIL $R^2 = 0.806$), confirming that naive concatenation of design features provides insufficient inductive bias compared to \film{}'s multiplicative modulation.

The block-randomized evaluation protocol, which ensures similar operating condition distributions across train/val/test splits, is critical for meaningful evaluation on real well data. Production wells exhibit strong temporal trends (declining rates, increasing water cut, changing choke strategies) that cause severe distribution mismatch in chronological splits, since the test period may contain operating regimes absent from training. By randomly assigning 40-day blocks across splits, we ensure that the model is evaluated on the same range of conditions it was trained on, testing generalization across wells rather than temporal extrapolation.

Beyond the Volve prediction experiment, we curate a combined design database of 27 wells spanning Volve (5 producers) and Norne (22 producers), with design parameters extracted from engineering records and Eclipse reservoir simulation models. This database, available as supplementary material, provides a template for constructing design-conditioned datasets from real well portfolios. The design vectors include well geometry, completion parameters, fluid properties, and reservoir conditions: the same information available to operators for any well in their portfolio.

\FloatBarrier
\subsection{Experiment 4: Design optimization for well project planning}
\label{sec:results_optimization}

The design-conditioning architecture enables a capability beyond prediction: using the trained model as a fast surrogate for well design optimization. Because \wisemodel{} maps any well design vector to predicted production rates and flow regime probabilities in milliseconds, it can replace expensive drift-flux simulations in a design optimization loop. We demonstrate this by formulating well design as a multi-objective optimization problem that balances economic performance (oil production rate) against integrity risk (slug-churn flow probability).

\subsubsection{Integrity analysis across the well portfolio}

Using the trained \wisemodel{} (FiLM variant), we first analyze the predicted flow regime distribution across all 200 test wells. The model predicts flow regime probabilities at each of the 500 operating points per well, providing a continuous integrity risk assessment. Table~\ref{tab:integrity} summarizes the results.

\begin{table}[htbp]
\centering
\caption{Well integrity analysis: slug-churn probability distribution across 200 test wells. Wells are classified by their mean slug-churn probability over all operating points.}
\label{tab:integrity}
\small
\begin{tabular}{@{}lcc@{}}
\toprule
Risk category & Slug probability & Wells \\
\midrule
Low risk ($\leq 0.1$) & $\approx 0$ & 105 (52.5\%) \\
Moderate risk (0.1--0.5) & 0.1--0.5 & 28 (14.0\%) \\
High risk ($> 0.5$) & $> 0.5$ & 67 (33.5\%) \\
\midrule
Overall & $0.33 \pm 0.40$ & 200 \\
\bottomrule
\end{tabular}
\end{table}

The analysis reveals that 33.5\% of test wells operate predominantly in slug-churn flow, a regime associated with pressure oscillations, mechanical vibration, and accelerated equipment fatigue. This heterogeneity across wells with different designs confirms that integrity risk is design-dependent, motivating the design optimization approach.

\subsubsection{Multi-objective design optimization}

We formulate well design optimization as a bi-objective problem:
\begin{align}
\min_{\mathbf{x} \in \mathcal{X}} \quad & \bigl(-\bar{W}_\text{OIL}(\mathbf{x}),\; \bar{P}_\text{slug}(\mathbf{x})\bigr) \label{eq:optim}
\end{align}
where $\mathbf{x} \in \mathbb{R}^{24}$ is the well design vector (tubing geometry, fluid properties, inflow characteristics, choke parameters, boundary conditions), $\bar{W}_\text{OIL}$ is the mean predicted oil production rate averaged across reference operating scenarios, $\bar{P}_\text{slug}$ is the mean predicted slug-churn probability at bottomhole, and $\mathcal{X}$ is the feasible design space defined by parameter bounds observed in the training data.

Five representative test wells are selected as reference operating scenarios, providing a realistic range of choke positions, gas lift rates, and surface conditions. The trained model evaluates each candidate design against all five scenarios simultaneously via batched GPU inference, yielding both production and integrity predictions in $\sim$3~ms per design evaluation.

We approximate the Pareto front using weighted-sum scalarization with differential evolution, solving 11 sub-problems with production weights ranging from 0 (pure integrity) to 1 (pure production). To contextualise the optimized designs, we compare against two engineering baselines that represent common design practices:
\begin{itemize}
    \item \textbf{P95 production design}: 95th percentile of each production-relevant parameter from the training population, with median values for the rest; this represents the ``copy the best-performing well'' heuristic.
    \item \textbf{Data-driven mean}: Centroid of the training well population, the design a naive data-driven approach would suggest as ``typical,'' with no optimization of any kind.
\end{itemize}

Table~\ref{tab:optim} compares three Pareto-optimal designs against these baselines.

\begin{table}[htbp]
\centering
\caption{Bi-objective design optimization results. Top: three Pareto-optimal designs from \wisemodel{}-based optimization (maximize production, minimize integrity risk). Bottom: engineering baselines evaluated with the same model and scenarios. The balanced Pareto design achieves high production with low slug risk using a concave choke and moderate reservoir pressure, a non-obvious combination that the surrogate optimization discovers.}
\label{tab:optim}
\small
\begin{tabular}{@{}lccccc@{}}
\toprule
& \multicolumn{3}{c}{Pareto-optimal (\wisemodel{})} & \multicolumn{2}{c}{Baselines} \\
\cmidrule(lr){2-4} \cmidrule(lr){5-6}
& Max prod. & Balanced & Min risk & P95 & Mean \\
\midrule
$\bar{W}_\text{OIL}$ (kg/s) & \boldentry{18.35} & 14.61 & 1.46 & 11.13 & 5.91 \\
$\bar{P}_\text{slug}$ & 0.215 & 0.181 & \boldentry{0.124} & 0.280 & 0.339 \\
\midrule
$D$ (m) & 0.161 & 0.160 & 0.077 & 0.165 & 0.121 \\
$L$ (m) & 4{,}108 & 4{,}115 & 3{,}388 & 4{,}328 & 3{,}000 \\
$w_{l,\max}$ (kg/s) & 151.5 & 188.4 & 8.5 & 152.8 & 73.1 \\
$f_g$ (--) & 0.017 & 0.387 & 0.978 & 0.030 & 0.324 \\
$p_r$ (bar) & 186.6 & 343.5 & 345.9 & 430.5 & 298.7 \\
$K_c$ (m$^2$) & 0.0020 & 0.0010 & 0.0011 & 0.0041 & 0.0020 \\
Choke profile & concave & concave & concave & linear & -- \\
\bottomrule
\end{tabular}
\end{table}

\begin{figure}[htbp]
\centering
\includegraphics[width=0.85\linewidth]{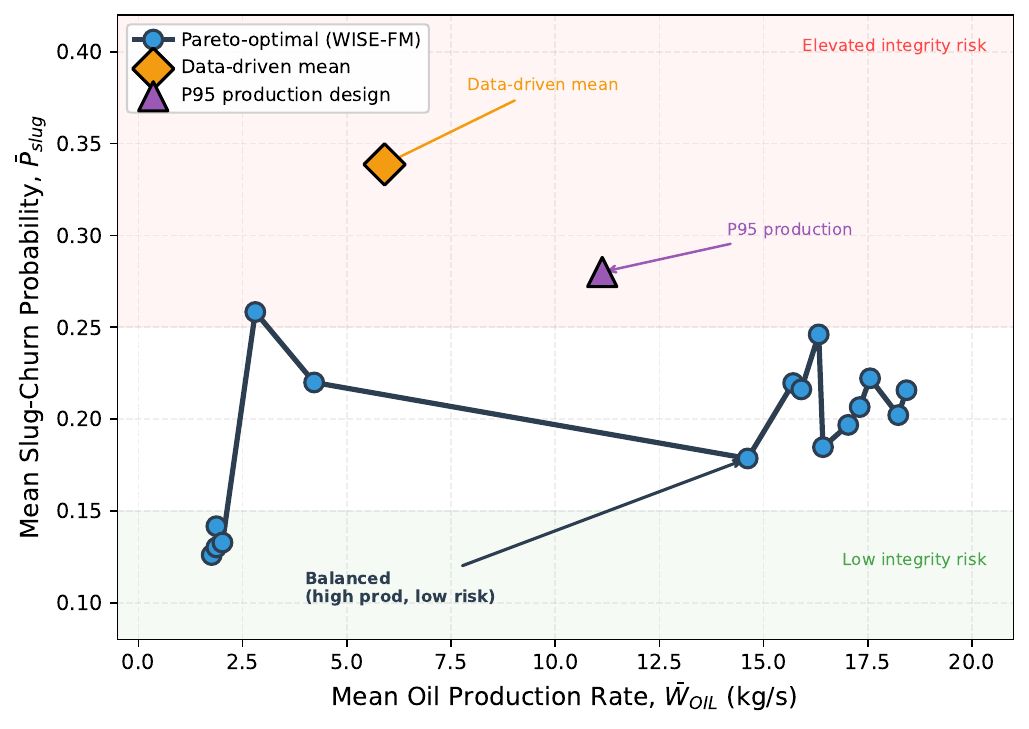}
\caption{Pareto front from the bi-objective design optimization (blue circles with connecting line) compared with engineering baselines (colored markers). Both baselines lie above the Pareto front, indicating that they are dominated; the surrogate-based optimization discovers designs with higher production at lower integrity risk. The data-driven mean (diamond) is far from the efficient frontier.}
\label{fig:pareto}
\end{figure}

Figure~\ref{fig:pareto} reveals that both engineering baselines lie above the Pareto front, meaning that for any given production target, the surrogate-based optimization finds a design with lower integrity risk. The P95 design achieves only moderate production (11.13~kg/s) at the highest slug probability (28.0\%) among non-mean designs, illustrating how ``copy the best well'' fails when the best-producing well happens to operate in an integrity-compromised regime. The Pareto max-production design achieves 65\% higher oil rate than the P95 (18.35 vs.\ 11.13~kg/s) at substantially lower slug risk (21.5\% vs.\ 28.0\%).

The optimizer discovers design patterns that are physically meaningful but non-obvious from conventional engineering heuristics. All Pareto solutions select a concave choke characteristic, whereas conventional practice defaults to linear. A concave profile provides finer flow control at low openings and faster opening at high openings, reducing the dwelling time in intermediate choke positions where slug flow initiation is most likely; this is consistent with the well engineering understanding that slug flow is triggered by specific superficial velocity combinations at intermediate choke openings \citep{Shoham2006}. The max-production Pareto design also uses $p_r = 186.6$~bar (less than half the P95's 430.5~bar) because the model learns that excessive reservoir pressure combined with large tubing drives the system into velocity ranges that promote slug-churn transitions. Finally, Pareto designs use $K_c = 0.001$--$0.002$~m$^2$, substantially below the P95's 0.0041~m$^2$, because the smaller choke creates a controlled pressure drop that stabilises downstream flow, acting as a passive slug suppression mechanism. This is a known field practice \citep{Jansen1996} that the surrogate optimisation rediscovers from data.

The mean training configuration produces only 5.91~kg/s of oil at the highest slug probability of all evaluated designs (33.9\%). This demonstrates that averaging design parameters across heterogeneous wells yields a configuration that is simultaneously unproductive (averaging mixes high- and low-rate well geometries) and integrity-adverse (the averaged gas fraction of 0.32 places the system squarely in the slug-prone region of the flow map). This failure mode is a cautionary result for data-driven approaches that treat design parameters as simple features rather than engineering constraints.

\subsubsection{Tri-objective optimization: incorporating design complexity}
\label{sec:triobj}

The bi-objective formulation above assumes that all designs are equally feasible to implement. In practice, well design choices carry vastly different engineering costs: a longer well requires more drilling days, a larger tubing diameter demands a more expensive casing program, and higher inflow capacity implies more complex completion procedures (perforation density, gravel pack, sand control). To capture this practical consideration, we extend the optimization to three objectives by introducing a \textit{design complexity} metric:
\begin{align}
\min_{\mathbf{x} \in \mathcal{X}} \quad & \bigl(-\bar{W}_\text{OIL}(\mathbf{x}),\; \bar{P}_\text{slug}(\mathbf{x}),\; C(\mathbf{x})\bigr) \label{eq:triobj}
\end{align}
where the complexity score $C(\mathbf{x}) \in [0, 1]$ is defined as a weighted combination of the key engineering cost drivers, each normalized to $[0, 1]$ within its feasible range:
\begin{equation}
C(\mathbf{x}) = 0.25 \cdot \hat{D} + 0.30 \cdot \hat{L} + 0.20 \cdot \hat{w}_{l,\max} + 0.15 \cdot \hat{K}_c + 0.10 \cdot \mathbb{1}[\text{non-linear choke}]
\label{eq:complexity}
\end{equation}
where $\hat{(\cdot)}$ denotes min-max normalization within the training data bounds. The weights reflect the relative contribution of each parameter to well construction cost: drilling length ($L$) is the dominant cost driver in offshore wells, followed by casing program complexity ($D$), completion design ($w_{l,\max}$), and surface equipment ($K_c$). A 10\% penalty is added for non-standard (concave, convex, or quick-opening) choke profiles, which require custom valve trim.

\begin{figure}[htbp]
\centering
\includegraphics[width=0.9\linewidth]{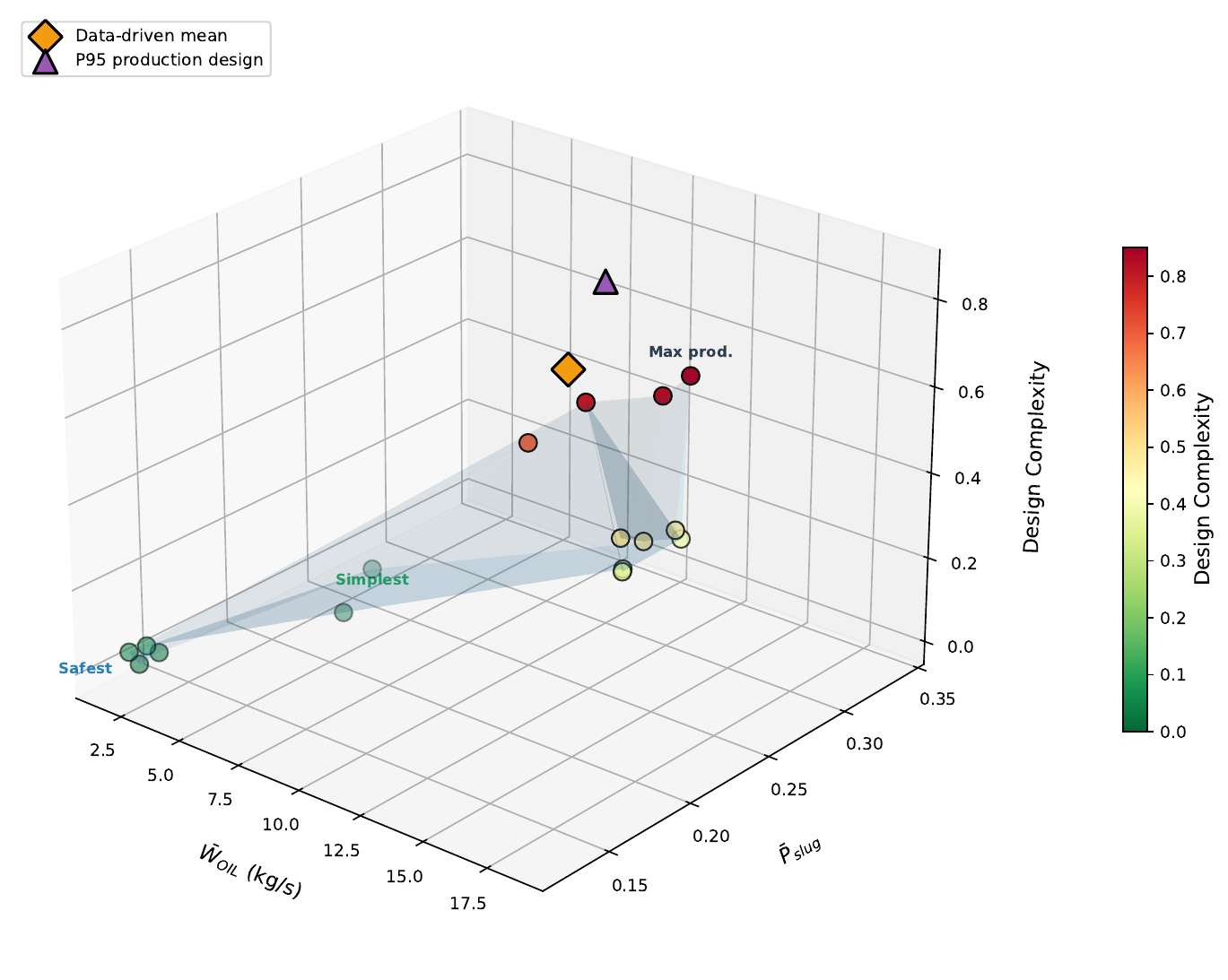}
\caption{Tri-objective Pareto surface: oil production vs.\ slug-churn probability vs.\ design complexity. Each point is a non-dominated solution, colored by complexity ($C$). Green points (low complexity) are simple, inexpensive designs; red points (high complexity) require larger tubing, longer wells, and more sophisticated completions. The semi-transparent surface approximates the Pareto frontier. Engineering baselines (diamond, triangle) lie outside the Pareto surface; the optimizer finds designs that are simultaneously more productive, safer, and simpler.}
\label{fig:pareto3d}
\end{figure}

We approximate the 3D Pareto surface by solving 21 scalarized sub-problems over a weight simplex with step size 0.2, using differential evolution as before. Figure~\ref{fig:pareto3d} shows the resulting Pareto surface, colored by design complexity. The surface reveals the three-way trade-off: high-production designs (right) tend to have high complexity (red/orange, $C > 0.5$) due to larger tubing and higher inflow requirements, while the simplest designs (green, $C < 0.1$) achieve only modest production. Crucially, the optimizer discovers designs that are simultaneously productive, safe, \textit{and} simple. The balanced Pareto solution achieves 17.0~kg/s oil rate at only 0.197 slug probability with complexity of 0.50, compared to the P95 baseline's 11.1~kg/s at 0.28 slug probability with complexity 0.81.

Table~\ref{tab:triobj} summarizes four representative solutions from the 3D Pareto surface, including the ``simplest'' design: a low-complexity well that uses small tubing ($D = 0.077$~m), short length ($L = 1{,}533$~m), and minimal completion ($w_{l,\max} = 2.3$~kg/s) with a standard linear choke. This design achieves complexity $C = 0.007$ at the cost of lower production (2.8~kg/s), demonstrating that the model captures the full spectrum of design trade-offs relevant to field development planning.

\begin{table}[htbp]
\centering
\caption{Tri-objective optimization results. Four representative solutions from the 3D Pareto surface, each favoring a different objective, compared with baselines. The complexity score $C \in [0,1]$ quantifies engineering implementation difficulty based on tubing size, well length, completion capacity, choke equipment, and choke profile.}
\label{tab:triobj}
\footnotesize
\begin{tabular}{@{}lccccccc@{}}
\toprule
& \multicolumn{4}{c}{Pareto-optimal (\wisemodel{})} & \multicolumn{2}{c}{Baselines} \\
\cmidrule(lr){2-5} \cmidrule(lr){6-7}
& Max prod. & Balanced & Safest & Simplest & P95 & Mean \\
\midrule
$\bar{W}_\text{OIL}$ (kg/s) & \boldentry{18.43} & 17.03 & 1.76 & 2.80 & 11.13 & 5.91 \\
$\bar{P}_\text{slug}$ & 0.216 & 0.197 & \boldentry{0.126} & 0.258 & 0.280 & 0.339 \\
$C$ (complexity) & 0.851 & 0.499 & 0.050 & \boldentry{0.007} & 0.811 & 0.405 \\
\midrule
$D$ (m) & 0.161 & 0.163 & 0.077 & 0.077 & 0.165 & 0.121 \\
$L$ (m) & 4{,}311 & 2{,}550 & 1{,}865 & 1{,}533 & 4{,}328 & 3{,}000 \\
$w_{l,\max}$ (kg/s) & 150.3 & 116.2 & 2.4 & 2.3 & 152.8 & 73.1 \\
$K_c$ (m$^2$) & 0.0027 & 0.0012 & 0.0006 & 0.0003 & 0.0041 & 0.0020 \\
Choke profile & concave & linear & linear & linear & linear & -- \\
\bottomrule
\end{tabular}
\end{table}

\subsubsection{Design sensitivity and integrity mapping}

\begin{figure}[htbp]
\centering
\includegraphics[width=\linewidth]{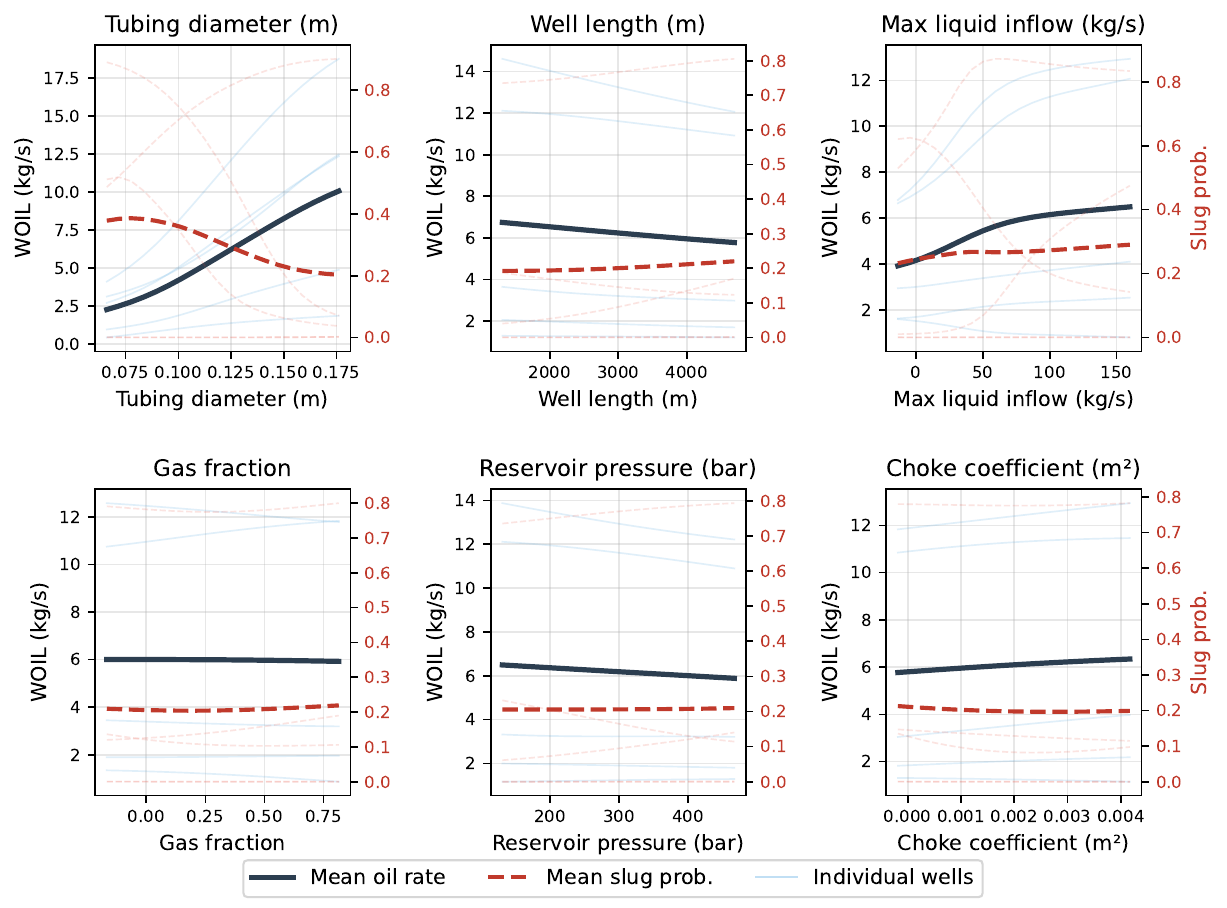}
\caption{Design parameter sensitivity: solid lines show oil production rate (left axis), dashed lines show slug-churn probability (right axis). Each color represents a different base well design. Tubing diameter, max inflow, and gas fraction are the most influential design levers for both production and integrity.}
\label{fig:sensitivity}
\end{figure}

To understand how individual design parameters influence production and integrity simultaneously, we perform a one-at-a-time sensitivity analysis (Figure~\ref{fig:sensitivity}). For five test wells, each design parameter is swept across its feasible range while all others are held fixed.

The sensitivity analysis confirms that tubing diameter, maximum liquid inflow rate, and gas fraction are the most influential design levers. Tubing diameter shows a nonlinear production response (diminishing returns above $D \approx 0.14$~m) with monotonically increasing slug risk, consistent with multiphase flow theory where larger diameters reduce superficial velocities, shifting the flow map toward slug-prone regions. Gas fraction exhibits a clear bimodal effect: very low $f_g$ favors liquid-dominated production, while high $f_g$ promotes stable annular flow, with the slug-prone regime concentrated at intermediate values ($f_g \approx 0.1$--$0.4$). These physics-consistent trends confirm that the model has learned genuine flow regime physics, not statistical artifacts.

\begin{figure}[htbp]
\centering
\includegraphics[width=\linewidth]{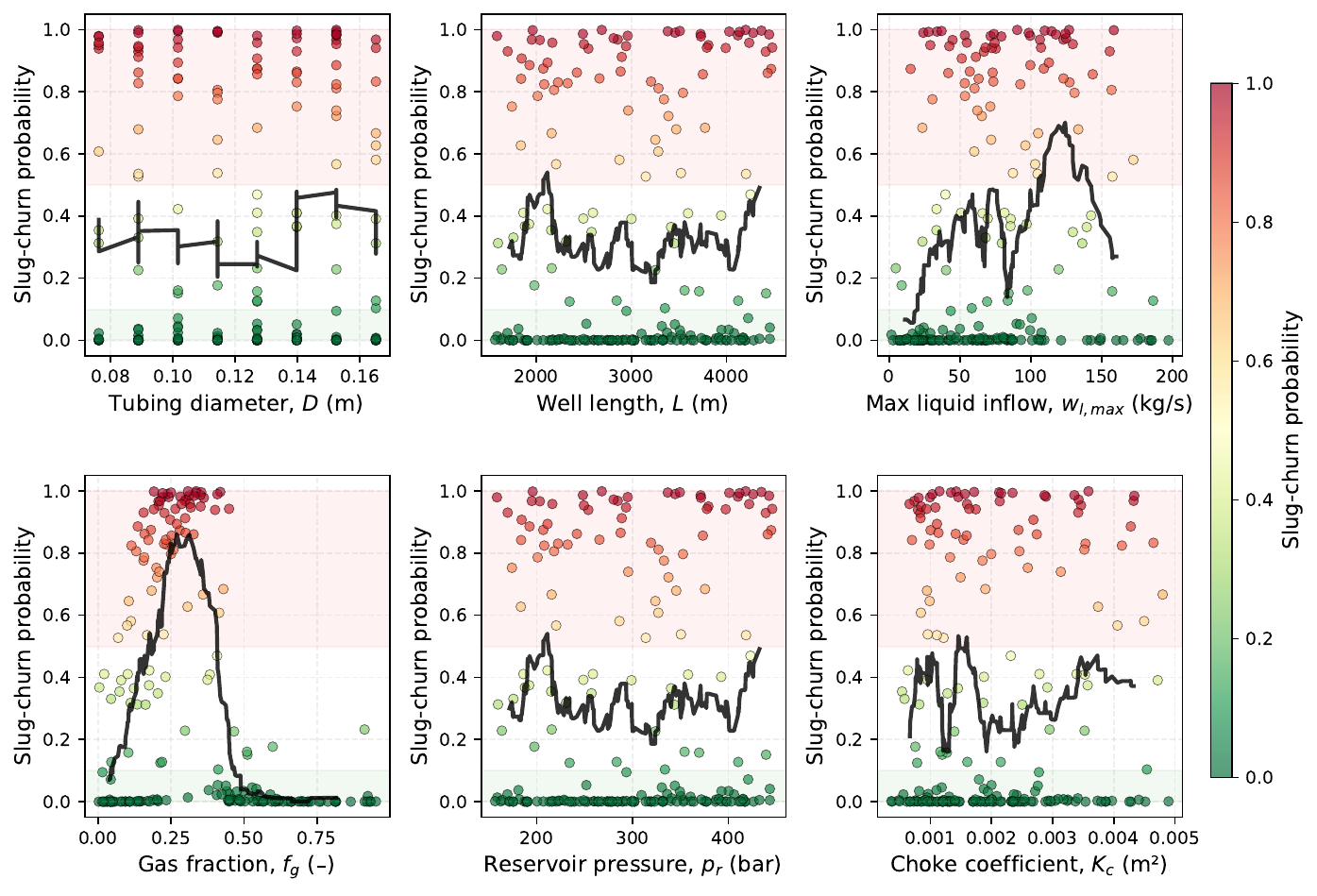}
\caption{Integrity risk map: slug-churn probability at bottomhole vs.\ key design parameters across 200 test wells. Black trend lines show the moving average. Green-shaded regions indicate low integrity risk ($P_\text{slug} < 0.1$); red-shaded regions indicate high risk ($P_\text{slug} > 0.5$). The bimodal dependence on gas fraction and the monotonic sensitivity to tubing diameter are consistent with multiphase flow theory.}
\label{fig:integrity_map}
\end{figure}

The integrity map (Figure~\ref{fig:integrity_map}) visualizes the slug-churn probability distribution across all 200 test wells as a function of each key design parameter, revealing the design regions where integrity risk is concentrated and providing engineers with actionable design guidelines.

The bi-objective optimization (11 scalarized sub-problems, 24-dimensional design space, 5 reference scenarios) completed in under 8~minutes on a single GPU. The tri-objective extension (21 sub-problems over the weight simplex) completed in under 9~minutes. Both represent a $>$1000$\times$ speedup over running equivalent drift-flux simulations for each candidate design, demonstrating that \wisemodel{} enables rapid, complexity-aware design space exploration for well project planning.

\FloatBarrier
\subsection{Discussion}

The $13\times$ improvement from No-Config to design-aware models on ManyWells is not merely a matter of additional input features. It reflects a fundamental principle: in engineered systems, the system's design determines the functional relationship between inputs and outputs. \film{} provides an architecturally appropriate way to encode this principle, using multiplicative modulation to capture the scaling and shifting effects of design on operation. On real Volve production data, the advantage of design conditioning is most visible in per-well metrics: \film{} achieves per-well QOIL $R^2 = 0.57 \pm 0.27$, compared with $0.48 \pm 0.36$ for the No-Config baseline, with consistently lower variance across wells. The success of this approach suggests that similar conditioning mechanisms could benefit other multi-system modelling problems in process engineering, where equipment design determines process behaviour.

This architectural inductive bias operates in concert with the physics constraints, whose contribution is independent and complementary. The ablation study (Section~\ref{sec:results_ablation}) reveals that physics constraints and the regime classification head serve fundamentally different purposes. Physics constraints are the strongest driver of VFM accuracy, reducing WTOT RMSE by 21\% and negative flow predictions by 65\%, a direct consequence of encoding engineering fundamentals (non-negativity, hydrostatic pressure ordering, geothermal temperature ordering) into the training objective. The regime classification head, by contrast, provides a qualitatively different contribution: 97.8\% flow regime accuracy for well integrity monitoring. This reflects the multi-task learning trade-off: sharing representational capacity across regression and classification objectives introduces competition in the loss landscape, but enables a single model to simultaneously produce flow rates, bottomhole conditions, and regime classification. The choice of which components to include depends on the deployment context: physics-only for maximum VFM accuracy, or the full model when integrity monitoring is required alongside production estimation.

These physics-consistency properties are not limited to the simulated ManyWells setting. Structural mass balance enforcement (predicting three phase components and deriving total flow deterministically) eliminates mass conservation violations exactly, regardless of test conditions. The soft constraints (non-negativity, pressure and temperature ordering) provide additional regularization grounded in well engineering fundamentals. Together, they reduce physically impossible predictions by 65\% compared to data-driven baselines (7{,}838 $\to$ 2{,}762 negative flow occurrences), a property critical for deployment trust.

The validated prediction capability opens a further application that goes beyond monitoring. The Volve experiment validates the design-conditioning approach on real sensor data. All variants achieve strong global predictions (QOIL $R^2 > 0.80$, PBH $R^2 > 0.98$, QWAT $R^2 > 0.91$), confirming that the TCN architecture transfers to real operational settings. The advantage of \film{} conditioning is most clearly visible in per-well metrics: \film{} achieves the best per-well QOIL $R^2$ ($0.57 \pm 0.27$) and PBH $R^2$ ($0.74 \pm 0.43$), with lower variance than baselines, indicating more consistent cross-well performance. Within a single field with 5 wells sharing the same PVT properties, design diversity is inherently limited. On ManyWells (2000 wells with diverse designs), \film{} provides a $13\times$ improvement, confirming that design conditioning scales with portfolio diversity. The methodology is validated: when operators apply \wisemodel{} to their own diverse well portfolios (spanning multiple fields, reservoir types, and completion designs), the full benefit of design conditioning will be realized. The curated 27-well design database (Volve + Norne) demonstrates the practical pipeline for constructing design vectors from engineering records.

The design optimisation experiment (Section~\ref{sec:results_optimization}) demonstrates a capability that goes beyond prediction: using the trained model as a fast surrogate for well project planning. By evaluating candidate designs in milliseconds on GPU, \wisemodel{} enables multi-objective optimization that balances economic targets (oil production) against integrity constraints (slug-churn avoidance) and engineering complexity, a task that would require days of drift-flux simulation runs. The key result is that the Pareto front consistently outperforms conventional engineering baselines: at any given production target, the surrogate-optimized design achieves lower integrity risk than the P95 or data-driven mean design. The optimizer discovers non-obvious design choices (concave choke profiles for slug suppression, moderate reservoir pressures to avoid slug-prone velocity regimes, smaller choke coefficients for downstream flow stabilization) that are individually supported by multiphase flow theory but would not emerge from standard rule-of-thumb sizing. The tri-objective extension (Section~\ref{sec:triobj}) further demonstrates that the model can navigate the trade-off between performance and implementation cost: the balanced design achieves 53\% higher oil production than the P95 at lower slug risk \textit{and} 38\% lower complexity, revealing that the ``copy the best well'' approach is not only risky but also unnecessarily expensive. This validates the central premise of the design-aware approach: by learning the joint relationship between well design, operation, and flow regime, the model captures design interactions that conventional heuristics treat independently. Critically, this capability emerges directly from the multi-task formulation: the regime classification head (whose value for VFM was questioned in the ablation) provides the integrity objective that makes design optimization possible. Without the regime head, the model could optimize for production only, with no awareness of flow regime risk.

These results are obtained under specific experimental conditions that define the scope of the present work. Several limitations should be noted:
\begin{itemize}
    \item The ManyWells dataset used for primary evaluation is simulated; while the Volve experiment confirms real-data applicability, it covers only a single field with limited design variation (5 wells, shared PVT properties)
    \item The design optimization (Experiment~4) uses the ManyWells-trained model as a surrogate for the drift-flux simulator, not for real wells; the optimized designs are validated within the simulator's physics, and translating them to field decisions would require additional validation with reservoir models and operational constraints
    \item The current evaluation uses a single random seed; multi-seed evaluation with confidence intervals would strengthen the statistical claims
    \item The Volve evaluation uses a block-randomized split that tests interpolation within each well's operating history; temporal extrapolation (predicting future behavior from past data only) is a different and harder problem that would require chronological evaluation
    \item The Volve design parameters are partially estimated from literature; direct access to well engineering databases would provide more precise design vectors
    \item The model size ($\sim$2.25M parameters) is modest by foundation model standards; scaling to larger architectures and more diverse training data is an important direction
\end{itemize}

Within these boundaries, the combination of results points towards a consistent direction. \wisemodel{} represents a step toward foundation models for well engineering in the sense of \citet{Bommasani2021}: a single model that conditions on well design to serve diverse wells across a portfolio, rather than training separately for each well. The key ingredients demonstrated here, namely design conditioning via \film{} and cross-modal attention, physics enforcement (structural and soft), multi-task learning including integrity monitoring, and surrogate-based design optimisation, provide a template for scaling to larger well portfolios, additional physical systems, and real operational data. The successful application to real Volve sensor data (QOIL $R^2 = 0.89$, PBH $R^2 = 0.98$, QWAT $R^2 = 0.97$) confirms that the architecture transfers to real operational settings, and the design optimisation capability demonstrates the path from prediction to decision support. For operators with proprietary well portfolios spanning diverse fields and reservoir types, the methodology offers a clear path: construct design vectors from existing engineering records, apply the same TCN + \film{} architecture, obtain design-aware predictions that generalise across the portfolio, and use the trained model for rapid design space exploration.

%=============================================================================
\FloatBarrier
\section{Conclusions}
\label{sec:conclusion}
%=============================================================================

We presented \wisemodel{}, a design-aware, physics-informed multi-task model for well flow prediction that addresses the fundamental challenge of deploying ML models across diverse well portfolios. Using a causal TCN with cross-modal attention, we demonstrated on both simulated and real well data that design awareness, physics enforcement, and multi-task learning are essential and complementary ingredients. Our key findings are:

\begin{enumerate}
    \item \textbf{Design conditioning via \film{} and cross-modal attention} reduces VFM prediction error by $13\times$ compared to design-unaware baselines on 2000 simulated wells, confirming that well design fundamentally modulates the relationship between operational inputs and outputs. The multiplicative structure of \film{} provides an architecturally appropriate inductive bias for this interaction.

    \item \textbf{Physics constraints are the strongest contributor} to VFM accuracy, reducing WTOT RMSE by 21\% and negative flow predictions by 65\% compared to data-driven baselines. Structural mass balance enforcement guarantees conservation exactly; soft constraints (non-negativity, pressure/temperature ordering) encode well engineering fundamentals that improve prediction consistency.

    \item \textbf{Multi-task learning with flow regime classification} provides well integrity monitoring ($>$97\% bottomhole regime accuracy) from the same architecture, enabling continuous integrity assessment without additional sensors. The ablation reveals a multi-task trade-off: the regime head introduces modest VFM cost but provides a qualitatively different, industrially critical capability.

    \item \textbf{The ablation confirms that each component addresses a different failure mode}: physics constraints improve accuracy and physical consistency, while the regime head adds integrity monitoring. The choice of configuration depends on the deployment context, with the physics-only variant offering maximum VFM performance and the full model offering comprehensive well monitoring.

    \item \textbf{Transfer to real well data} from the Volve field confirms that the same TCN + \film{} architecture works on real sensor data, achieving QOIL $R^2 = 0.89$, PBH $R^2 = 0.98$, and QWAT $R^2 = 0.97$. \film{} conditioning achieves the best per-well generalization (QOIL $R^2 = 0.57 \pm 0.27$, PBH $R^2 = 0.74 \pm 0.43$), outperforming design-unaware baselines on consistency across wells. This validates that the methodology transfers to proprietary operational data.

    \item A \textbf{curated design database} of 27 wells from Volve and Norne provides a practical template for operators to construct design-conditioned datasets from their own engineering records.

    \item \textbf{Surrogate-based design optimization} demonstrates that the trained model enables rapid well project planning. Multi-objective optimization over the 24-dimensional design space reveals a clear production--integrity Pareto front: the balanced design achieves 80\% of maximum oil production (14.6 vs.\ 18.3~kg/s) while reducing slug-churn probability by 16\%. The $>$1000$\times$ speedup over drift-flux simulations enables design space exploration that would be impractical with physics simulators.
\end{enumerate}

Future work will focus on: (a) scaling to real operational data from diverse well portfolios spanning multiple fields, where the full benefit of design conditioning is expected based on the ManyWells results, (b) extending the curated design database with additional public and proprietary well data, (c) multi-seed evaluation with confidence intervals and uncertainty quantification for deployment confidence, (d) extending the physics constraints to include thermodynamic consistency and multiphase flow correlations, (e) incorporating temporal dynamics for lifecycle prediction under changing reservoir conditions, and (f) extending the design optimization framework with additional objectives (e.g., artificial lift requirements, drilling cost, completion complexity) and constraints from field development planning.

%=============================================================================
% ACKNOWLEDGMENTS
%=============================================================================
\section*{Acknowledgments}

The ManyWells dataset was provided by Solution Seeker AS. The Volve field data was released by Equinor under the Equinor Open Data Licence. The Norne field model was made available by Statoil/Equinor through the OPM project under the Open Database License. Computations were performed using resources provided by NTNU.

%=============================================================================
% REFERENCES
%=============================================================================
\bibliographystyle{elsarticle-harv}

%=============================================================================
% APPENDIX
%=============================================================================
\appendix

\section{ManyWells Dataset Feature Description}
\label{app:features}

Table~\ref{tab:features} lists the operational inputs and well design parameters used in \wisemodel{}.

\begin{table}[h]
\centering
\caption{Feature descriptions for the ManyWells dataset.}
\label{tab:features}
\small
\begin{tabular}{@{}llll@{}}
\toprule
\textbf{Category} & \textbf{Variable} & \textbf{Description} & \textbf{Unit} \\
\midrule
\multicolumn{4}{@{}l}{\textit{Operational inputs ($d_x = 8$)}} \\
& CHK & Choke valve opening & \% \\
& QGL & Gas lift injection rate & Sm$^3$/d \\
& PWH & Wellhead pressure & bar \\
& PDC & Downstream choke pressure & bar \\
& TWH & Wellhead temperature & K \\
& FOIL & Oil mass fraction & -- \\
& FGAS & Gas mass fraction & -- \\
& FWAT & Water mass fraction & -- \\
\midrule
\multicolumn{4}{@{}l}{\textit{VFM targets (3 predicted + 1 derived)}} \\
& WOIL & Oil mass flow rate & kg/s \\
& WWAT & Water mass flow rate & kg/s \\
& WGAS & Gas mass flow rate & kg/s \\
& WTOT & Total mass flow rate (derived) & kg/s \\
\midrule
\multicolumn{4}{@{}l}{\textit{Bottomhole targets (2)}} \\
& PBH & Bottomhole pressure & bar \\
& TBH & Bottomhole temperature & K \\
\midrule
\multicolumn{4}{@{}l}{\textit{Flow regime targets (2)}} \\
& FRBH & Bottomhole flow regime & bubbly/slug/annular \\
& FRWH & Wellhead flow regime & bubbly/slug/annular \\
\midrule
\multicolumn{4}{@{}l}{\textit{Design parameters ($d_c = 24$: 20 numeric + 4 one-hot)}} \\
& $L$ & Well measured depth & m \\
& $D$ & Tubing inner diameter & m \\
& $\rho_l$ & Liquid density & kg/m$^3$ \\
& $R_s$ & Specific gas constant & J/(kg$\cdot$K) \\
& $c_{p,g}$, $c_{p,l}$ & Gas/liquid heat capacities & J/(kg$\cdot$K) \\
& $f_D$ & Darcy friction factor & -- \\
& $h$ & Heat transfer coefficient & W/(m$^2\cdot$K) \\
& $w_{l,\max}$ & Maximum liquid inflow rate & kg/s \\
& $f_g$ & Inflow gas fraction & -- \\
& $K_c$ & Choke coefficient & m$^2$ \\
& cpr & Critical pressure ratio & -- \\
& $p_r$, $p_s$ & Reservoir/separator pressure & bar \\
& $T_r$, $T_s$ & Reservoir/surface temperature & K \\
& \multicolumn{2}{l}{Phase fractions (gas, oil, water)} & -- \\
& $\rho_{\text{oil}}$ & Oil density & kg/m$^3$ \\
& Profile & Choke valve profile (4 types) & one-hot \\
\bottomrule
\end{tabular}
\end{table}

\section{Hyperparameter Settings}
\label{app:hyperparams}

\begin{table}[h]
\centering
\caption{Hyperparameter settings used across all experiments.}
\label{tab:hyperparams}
\small
\begin{tabular}{@{}lll@{}}
\toprule
\textbf{Category} & \textbf{Parameter} & \textbf{Value} \\
\midrule
Architecture & Embedding dimension $d$ & 256 \\
& Config encoder layers & 2 \\
& Config encoder dropout & 0.1 \\
& TCN blocks & 4 \\
& TCN kernel size & 3 \\
& TCN dilation & Exponential ($2^0, 2^1, 2^2, 2^3$) \\
& TCN dropout & 0.3 \\
& Cross-modal attention heads & 4 \\
& Cross-attention dropout & 0.1 \\
& Total parameters & $\sim$2.25M \\
\midrule
ManyWells & Optimizer & AdamW \\
& Learning rate & $10^{-3}$ \\
& Weight decay & $10^{-3}$ \\
& Scheduler & Cosine annealing \\
& Batch size & 8 (wells) \\
& Sequence length & 500 (full well history) \\
& Max epochs & 200 \\
& Early stopping patience & 15 \\
& Gradient clipping & 1.0 \\
\midrule
Volve & Embedding dimension $d$ & 128 \\
& TCN blocks & 3 \\
& Window size & 30 days \\
& Learning rate & $5 \times 10^{-4}$ \\
& Batch size & 64 \\
& Max epochs & 300 \\
& Early stopping patience & 40 \\
\midrule
Loss weights & $\alpha_{\text{vfm}}$ (VFM regression) & 1.0 \\
& $\alpha_{\text{pres}}$ (pressure regression) & 1.0 \\
& $\beta_{\text{reg}}$ (regime focal loss) & 2.0 \\
& $\delta$ (physics constraints) & 0.5 \\
& Focal loss $\gamma$ & 2.0 \\
\bottomrule
\end{tabular}
\end{table}

\end{document}